\documentclass[lettersize,journal]{IEEEtran}

\usepackage{amsmath,amsfonts,amssymb}
\usepackage{algorithmic}
\usepackage{algorithm}
\usepackage{array}
\usepackage[caption=false,font=footnotesize,labelfont=rm,textfont=rm]{subfig}
\usepackage{textcomp}
\usepackage{stfloats}
\usepackage{url}
\usepackage{graphicx}
\usepackage{cite}
\usepackage{bm}
\usepackage{booktabs}
\usepackage{multirow}
\usepackage{mathtools}

\newcommand{\R}{\mathbb{R}}
\newcommand{\SO}{\mathrm{SO}}
\newcommand{\so}{\mathfrak{so}}
\newcommand{\SE}{\mathrm{SE}}
\newcommand{\se}{\mathfrak{se}}
\newcommand{\dd}{\mathrm{d}}
\newcommand{\T}{\top}
\newcommand{\eye}{\mathbf{I}}
\newcommand{\zero}{\mathbf{0}}
\newcommand{\skewop}[1]{\left[#1\right]^\wedge}
\DeclareMathOperator{\diag}{diag}

\begin{document}

\title{WNOJ-LIO: A White-Noise-on-Jerk Motion-Prior EKF\\for High-Dynamic LiDAR-IMU Fusion}

\author{Junning~Lyu,~Qizhi~Guo,~Xia~Ning,~Tao~Song,~and~Shaoming~He%
\thanks{Junning Lyu, Qizhi Guo, Xia Ning and Tao Song are with the School of Aerospace Engineering, Beijing Institute of Technology, Beijing 100081, China.}%
\thanks{Shaoming He (corresponding author) is with the School of Aerospace Engineering, Beijing Institute of Technology, Beijing 100081, China. (e-mail: heshaoming@bit.edu.cn)}%
\thanks{This work was supported by the National Key Research and Development Program Grant No. 2022YFE0204400.}}

\markboth{Journal of \LaTeX\ Class Files,~Vol.~14, No.~8, August~2021}{Lyu \MakeLowercase{\textit{et al.}}: WNOJ-LIO for High-Dynamic LiDAR-IMU Fusion}

\maketitle

\begin{abstract}
    LiDAR-inertial odometry (LIO) is a key component of autonomous navigation, but high-dynamic driving exposes two coupled challenges: intra-scan motion distortion and vibration-contaminated inertial measurements. Most real-time LiDAR-inertial pipelines propagate the system state by integrating raw IMU measurements and then use the propagated trajectory for point cloud de-distortion, thereby propagating inertial noise into both the corrected scan and the subsequent scan-to-map registration. This paper presents WNOJ-LIO, a LiDAR-IMU fusion framework based on a White-Noise-on-Jerk (WNOJ) Extended Kalman Filter (EKF). WNOJ-LIO employs a decoupled WNOJ prior on $\R^3 \times \SO(3)$ for state prediction and treats the IMU as a high-frequency measurement source rather than the driver of state propagation. The resulting posterior state history is then used for LiDAR scan de-distortion and subsequent point-to-plane LiDAR updates. The decoupled process model enables closed-form covariance propagation, thereby bridging the gap between batch WNOJ Gaussian process (GP) trajectory priors and recursive filtering.
    Simulation results demonstrate improvements in acceleration and angular-velocity denoising, scan de-distortion, and localization accuracy over a FAST-LIO-style baseline. 
    Real-world experiments were conducted using an autonomous racing car on four driving segments with maximum speeds ranging from 53 to 208~km/h, covering a wide range of vehicle vibration levels. The experiments further validate the proposed method and provide a comprehensive evaluation of its performance in estimating acceleration, angular velocity, body-frame linear velocity, attitude, and position under highly dynamic driving. The source code of WNOJ-LIO is publicly available at \url{https://github.com/LvJohny/wnoj-ekf-lio.git}.
\end{abstract}


\section{Introduction}
\label{sec:introduction}

\IEEEPARstart{A}{ccurate} and real-time state estimation is fundamental to autonomous systems operating under highly dynamic conditions, such as autonomous racing~\cite{pedoneRacecar2020,maDynamic2026}, aggressive autonomous driving~\cite{tangEnhanced2025,wangNovel2026}, and high-speed aerial navigation~\cite{yangTightly2025}. Modern LiDAR-inertial odometry has become the dominant solution for this task by combining the geometric accuracy of LiDAR \cite{zhangLOAM2014} with the high temporal resolution of inertial measurements~\cite{zhangLILO2022,quLLOL2022,qinLINS2020}. 

In most existing LIO systems, high-rate IMU measurements are first integrated to propagate the platform trajectory, and the resulting trajectory is subsequently used to de-distort each LiDAR scan before scan-to-map registration~\cite{xuFASTLIO22022,shanLIOSAM2020}. Although this tightly coupled pipeline is computationally efficient and performs well in ordinary driving scenarios, it also causes inertial measurement noise to be directly propagated into the corrected point cloud. As a result, the quality of LiDAR registration becomes inherently dependent on the quality of IMU propagation. This issue becomes particularly pronounced during highly dynamic maneuvers, where severe vehicle vibration contaminates inertial measurements~\cite{capriglionePerformance2021, xuAdaptive2023} while rapid platform motion simultaneously demands highly accurate scan de-distortion.

A direct way to alleviate this coupling is to avoid relying on IMU-driven trajectory propagation and instead reconstruct the scan trajectory using an explicit motion model. Early approaches~\cite{zhangLOAM2014, bosseContinuous2009, wangFLOAM2021} employ low-order kinematic assumptions, typically constant linear/angular velocity~\cite{wangFLOAM2021}, to predict the sensor motion during a scan. More recent continuous-time methods adopt higher-order motion priors to obtain more expressive trajectory representations, including GP models based on the WNOJ prior~\cite{tangWhiteNoiseonJerk2019}, the Singer motion model~\cite{wongDataDriven2020}, and hybrid formulations that assume constant acceleration for translation and constant angular velocity for rotation~\cite{shenCTEMLO2025}. These approaches improve trajectory reconstruction without relying on raw IMU integration. However, relying solely on a motion prior inevitably sacrifices the high-frequency motion constraints provided by inertial measurements. 

Rather than replacing IMU measurements entirely, a more attractive solution is to combine expressive motion priors with high-frequency inertial constraints. Recent continuous-time LiDAR-inertial systems have demonstrated that GP motion priors provide smooth trajectory interpolation, improved motion compensation, and principled IMU preintegration~\cite{burnettContinuousTime2025,zhengTrajLIO2024}. Meanwhile, recursive EKF-based methods, represented by FAST-LIO and its variants, have achieved remarkable accuracy, robustness, and computational efficiency for real-time LiDAR-IMU fusion, establishing the EKF as the dominant front-end architecture in modern LIO systems. This naturally motivates combining higher-order motion priors with recursive EKF estimation, allowing the motion model to provide physically consistent trajectory prediction while retaining the high-rate correction capability of inertial measurements.

Despite this appealing perspective, existing higher-order motion priors are almost exclusively formulated for continuous-time optimization rather than recursive Bayesian filtering. In particular, the classical continuous-time model, such as WNOJ, represents the system state on $\SE(3)$, where translational and rotational motions are tightly coupled and the associated covariance is defined only in the local pose space. These formulations are therefore incompatible with the recursive covariance propagation and state-update mechanism of the EKF, leaving the integration of higher-order motion priors into EKF-based LiDAR-IMU fusion largely unexplored.

\begin{figure*}[t]
    \centering
    \includegraphics[width=\textwidth]{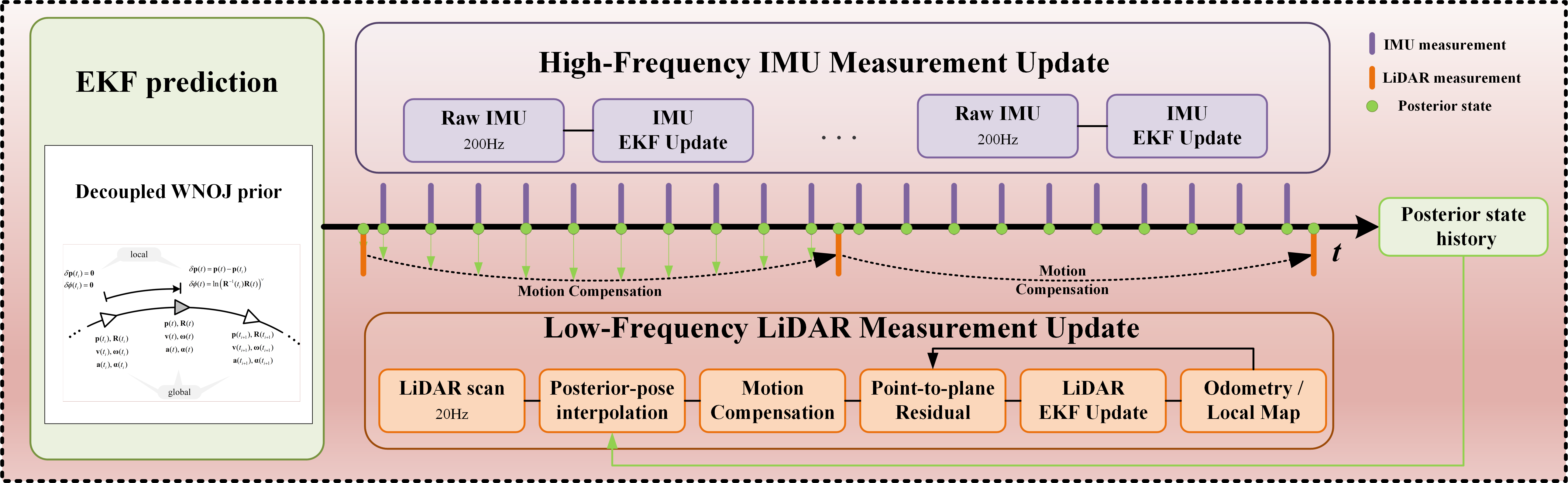}
    \caption{Overview of the WNOJ-LIO framework. The decoupled WNOJ prior drives EKF prediction, high-frequency IMU measurements update the state at 200~Hz, and low-frequency LiDAR measurements at 20~Hz use posterior-pose interpolation for motion compensation.}
    \label{fig:wnoj_lio_framework}
\end{figure*}

To address this limitation, this paper proposes WNOJ-LIO, a LiDAR-IMU fusion framework based on a WNOJ EKF for high-dynamic state estimation. Instead of using the IMU as the driver of state propagation, the proposed framework employs a decoupled WNOJ prior on $\mathbb{R}^3 \times \mathrm{SO}(3)$ to predict the system state, while treating IMU measurements as high-frequency EKF updates. The posterior state history generated by the recursive IMU updates is subsequently used for LiDAR scan de-distortion and point-to-plane registration. This architecture enables the motion prior to suppress vibration-induced inertial disturbances while allowing IMU measurements to continuously correct the predicted trajectory, thereby combining the complementary strengths of model-based prediction and measurement-driven estimation. The resulting process model admits closed-form covariance propagation, providing a principled bridge between continuous-time GP motion priors and recursive EKF-based LiDAR-IMU fusion. The overall framework is illustrated in Fig.~\ref{fig:wnoj_lio_framework}.

The main contributions of this paper are:
\begin{enumerate}
    \item A decoupled WNOJ process model on $\mathbb{R}^3 \times \mathrm{SO}(3)$ that enables recursive EKF filtering through closed-form state and covariance propagation, thereby making higher-order GP motion priors compatible with recursive LiDAR-IMU fusion.
    \item A dual-frequency LiDAR-IMU fusion architecture that treats IMU measurements as EKF updates rather than the prediction driver, reducing the propagation of vibration-induced inertial disturbances into LiDAR scan de-distortion through a higher-order WNOJ motion prior.
    \item Extensive simulation and real-world experiments on autonomous racing demonstrate improved inertial-state estimation, body-frame velocity, attitude estimation, and competitive localization under highly dynamic motion.
\end{enumerate}

The remainder of this paper is organized as follows. Section~\ref{sec:related_work} reviews related work. Section~\ref{sec:preliminaries} formulates the problem. Section~\ref{sec:decoupled_wnoj} presents the decoupled WNOJ process model, including local variables, global-state prediction, and covariance propagation. Section~\ref{sec:ekf_design} instantiates this process model in a WNOJ-LIO-based EKF with augmented states, IMU update, de-distortionand, and LiDAR updates. Sections~\ref{sec:simulation} and~\ref{sec:experiments} present simulation and racing-car experiments, respectively. Section~\ref{sec:conclusion} concludes the paper and summarizes limitations.

\section{Related Work}
\label{sec:related_work}
\subsection{Continuous-Time State Estimation and WNOJ Motion Priors}

Continuous-time state estimation provides a natural framework for representing platform motion during a LiDAR scan and for fusing asynchronous sensor measurements. Rather than estimating a single pose for each scan, it models the platform trajectory as a continuous function of time, allowing poses to be queried at arbitrary point timestamps. Existing continuous-time methods can generally be divided into parametric and nonparametric trajectory representations. Parametric approaches typically represent the trajectory using B-splines~\cite{nguyenSLICT2023,lvContinuousTime2023,quenzelRealtime2021,lvCLINS2021,langCocoLIC2023,sommerEfficient2020} or piecewise linear interpolation~\cite{dellenbachCTICP2022,zhengTrajLO2024,zhouATICTLO2024}. In contrast, nonparametric approaches formulate the trajectory as a Gaussian process, leading to exactly sparse continuous-time estimation on $\SE(3)$~\cite{andersonFull2015,tangWhiteNoiseonJerk2019,wongDataDriven2020} and its extensions to matrix Lie groups~\cite{dongSparse2018}. These formulations have demonstrated the advantages of continuous-time trajectories for motion compensation and asynchronous multi-sensor fusion. 

Linear interpolation offers the lowest computational cost but relies on piecewise constant-velocity assumptions, which become less accurate during aggressive maneuvers. To improve trajectory fidelity, several methods increase the density of trajectory control poses~\cite{zhengTrajLO2024} or adopt higher-order spline representations~\cite{langCocoLIC2023}. 

In contrast, GP-based formulations explicitly model system kinematics through stochastic differential equations, producing trajectory priors whose state variables correspond directly to the kinematic quantities required for autonomous estimation and control.
Within GP-based continuous-time estimation, the choice of motion prior determines both the system state and the assumed trajectory smoothness. The White-Noise-on-Acceleration (WNOA) prior models constant generalized acceleration perturbed by white noise~\cite{andersonFull2015}, whereas the WNOJ prior extends the system to a third-order stochastic process on local pose by explicitly modeling jerk~\cite{tangWhiteNoiseonJerk2019}. Data-driven Singer priors further learn motion statistics from trajectory datasets~\cite{wongDataDriven2020}. However, existing GP-prior-based LiDAR-inertial systems~\cite{wuPicking2023,zhengTrajLIO2024} do not directly offer the same prediction-update structure and runtime profile as EKF-based LIO.

\subsection{LiDAR-Inertial Fusion}

Although LiDAR-only odometry has achieved remarkable performance, high-rate inertial measurements are widely incorporated to improve state estimation and provide motion information for LiDAR scan de-distortion. Existing LiDAR--inertial odometry systems can be broadly categorized into optimization-based and filtering-based frameworks according to how IMU measurements are processed.

Optimization-based methods, such as LIO-SAM~\cite{shanLIOSAM2020} and LIC-Fusion~\cite{zuoLICFusion2020}, typically process IMU measurements in two stages. First, raw inertial measurements are propagated forward to provide an initial trajectory for scan de-distortion and optimization initialization. Subsequently, IMU preintegration is formulated as a factor to constrain the optimized trajectory within a graph or sliding window. Continuous-time optimization methods, including CLINS, SLICT, and Coco-LIC~\cite{lvCLINS2021,nguyenSLICT2023,langCocoLIC2023}, follow a similar strategy while replacing the discrete trajectory with a continuous-time representation and introducing continuous-time motion priors into the optimization.

Filtering-based methods provide a different trade-off by maintaining a fixed-size state and processing measurements recursively. Representative systems such as FAST-LIO and FAST-LIO2 tightly couple IMU propagation with LiDAR residuals within an iterated Extended Kalman Filter (IEKF), achieving excellent computational efficiency and real-time performance~\cite{xuFASTLIO2021,xuFASTLIO22022}. In these methods, however, the IMU serves as the process model driving state prediction, and the propagated trajectory is directly used for scan de-distortion.

More recently, continuous-time LiDAR--inertial systems have explored an alternative formulation in which IMU measurements are treated as observations of the continuous-time trajectory rather than as inputs to the process model~\cite{zhengTrajLIO2024,burnettIMU2025}. This formulation enables the motion prior to estimate the entire trajectory while allowing inertial measurements to constrain the underlying kinematic states. Nevertheless, these methods rely on sliding-window optimization, where a large number of trajectory states associated with high-rate IMU measurements must be optimized simultaneously, resulting in substantially higher computational cost than EKF-based methods. Consequently, integrating continuous-time motion priors with the computational efficiency of recursive EKF-based LiDAR-IMU fusion remains largely unexplored.

\begin{table}[hbtp]
\caption{Comparison of representative LiDAR--inertial odometry methods.}
\label{tab:method_comparison}
\centering
\scriptsize
\renewcommand{\arraystretch}{1.15}
\setlength{\tabcolsep}{3pt}
\begin{tabular*}{\columnwidth}{@{\extracolsep{\fill}}lcccc@{}}
\toprule
\textbf{System} & \textbf{\shortstack{Continuous-Time\\Estimation}} & \textbf{\shortstack{Prediction\\Model}} & \textbf{\shortstack{IMU\\Role}} & \textbf{\shortstack{Estimator\\Type}} \\
\midrule
LOAM~\cite{zhangLOAM2014} & $\times$ & Constant velocity & None & Optimization \\
LIO-SAM~\cite{shanLIOSAM2020} & $\times$ & IMU propagation &  Preintegration & Optimization \\
FAST-LIO~\cite{xuFASTLIO2021} & $\times$ & IMU propagation & Prediction & IEKF \\
Traj-LIO~\cite{zhengTrajLIO2024} & $\checkmark$ & GP motion prior & Measurement & Optimization \\
\midrule
\textbf{Ours} & $\mathbf{\checkmark}$ & \textbf{GP motion prior} & \textbf{Measurement} & \textbf{EKF} \\
\bottomrule
\end{tabular*}
\end{table}


\section{Problem Formulation and Preliminaries}
\label{sec:preliminaries}

\subsection{Notation and Coordinate Conventions}

We describe platform motion with position, velocity, acceleration, rotation, angular velocity, and angular acceleration. These quantities are denoted by $\mathbf{p}$, $\mathbf{v}$, $\mathbf{a}$, $\mathbf{R}$, $\bm{\omega}$, and $\bm{\alpha}$ when no coordinate-frame ambiguity exists. Two coordinate frames are used throughout the paper: the world frame $w$ and the platform body frame $b$. A superscript indicates the frame in which a vector is expressed, and a two-letter subscript indicates the physical quantity or transformation from the second frame to the first frame. For example, $\mathbf{p}^w_{wb}$ is the position of the body frame relative to the world frame, expressed in $w$, and $\mathbf{R}_{wb}$ maps body-frame vectors into the world frame. In the LiDAR-IMU fusion context, the body frame coincides with the LiDAR frame, and the IMU frame is related via a known extrinsic calibration.

We use standard Lie-group notation for rotations and rigid-body poses. The rotation matrix satisfies $\mathbf{R}\in\SO(3)$, where $\SO(3)$ is the three-dimensional rotation group, and $\so(3)$ denotes its Lie algebra. The rigid-body pose is represented by $\mathbf{T}=\begin{bmatrix}\mathbf{R}&\mathbf{p}\\ \zero&1\end{bmatrix}\in\SE(3)$, where $\SE(3)$ is the special Euclidean group and $\se(3)$ is its Lie algebra. A local pose perturbation on $\SE(3)$ is denoted by $\bm{\xi}=[\bm{\rho}^\T,\bm{\phi}^\T]^\T\in\R^6$, where $\bm{\rho}$ is the translational perturbation and $\bm{\phi}$ is the rotational perturbation. The hat operator $\skewop{\cdot}$ maps a vector to a skew-symmetric matrix. The operators $\exp(\cdot)$ and $\log(\cdot)$ are used for both $\SO(3)$ and $\SE(3)$. For rotations, $\exp: \so(3) \to \SO(3)$ recovers a rotation matrix from its Lie algebra element, and $\log: \SO(3) \to \so(3)$ is the inverse. The right Jacobian $\mathbf{J}_r(\bm{\phi})$ and its inverse satisfy the first-order approximations~\cite{solaMicro2021}
\begin{equation}
    \mathbf{J}_r(\bm{\phi}) \approx \eye - \tfrac{1}{2}\skewop{\bm{\phi}}, \quad
    \mathbf{J}_r^{-1}(\bm{\phi}) \approx \eye + \tfrac{1}{2}\skewop{\bm{\phi}},
    \label{eq:jr_approx}
\end{equation}
valid for small $\|\bm{\phi}\|$.

\subsection{Gaussian Process Kinematic Priors}

Many continuous-time trajectory estimation methods employ GP priors represented by linear time-invariant (LTI) stochastic differential equations (SDEs)~\cite{tangWhiteNoiseonJerk2019}
\begin{equation}
    \dot{\bm{\gamma}}(t) = \mathbf{A}\,\bm{\gamma}(t) + \mathbf{L}\,\mathbf{w}(t), \quad
    \mathbf{w}(t) \sim \mathcal{GP}(\zero, \mathbf{Q}_c\,\delta(t-\tau)),
    \label{eq:gp_sde}
\end{equation}
where $\bm{\gamma}(t)$ is the state, $\mathbf{w}(t)$ is a white-noise GP with power spectral density $\mathbf{Q}_c$, and $\delta(\cdot)$ is the Dirac delta. The prior mean propagates via the state transition function $\bm{\Phi}$ according to
\begin{equation}
    \hat{\bm{\gamma}}(\tau) = \bm{\Phi}(\tau, t_k)\,\hat{\bm{\gamma}}(t_k), \quad
    \bm{\Phi}(\tau, t_k) = \exp(\mathbf{A}\,\Delta t),
\end{equation}
where $\Delta t = \tau - t_k$.

\subsection{Classical WNOJ on $\SE(3)$ and Its Limitations}
\label{sec:classical_wnoj}

The classical WNOJ model~\cite{tangWhiteNoiseonJerk2019} represents the trajectory on $\SE(3)$ as
\begin{equation}
    \mathbf{T}(t) = \exp(\bm{\xi}^\wedge) =
    \begin{bmatrix} \mathbf{R}(t) & \mathbf{p}(t) \\ \zero & 1 \end{bmatrix} \in \SE(3),
\end{equation}
with the global state $\{\mathbf{T}(t), \bm{\varpi}(t), \dot{\bm{\varpi}}(t)\}$, where $\bm{\varpi}(t) \in \R^6$ is the body-frame generalized velocity and $\dot{\bm{\varpi}}(t) \in \R^6$ is the body-frame generalized acceleration. Since the pose kinematics $\dot{\mathbf{T}} = \bm{\varpi}^\wedge\mathbf{T}$ is nonlinear, we define a \emph{local pose variable} as
\begin{equation}
    \delta\bm{\xi}_i(t) := \log\!\left(\mathbf{T}(t)\,\mathbf{T}_i^{-1}\right), \quad t \in [t_i, t_{i+1}),
\end{equation}
where $\mathbf{T}_i = \mathbf{T}(t_i)$. The augmented local state
$\bm{\gamma}_i(t) = {[\delta\bm{\xi}_i^\T,\; \dot{\delta\bm{\xi}}_i^\T,\; \ddot{\delta\bm{\xi}}_i^\T]}^\T$
satisfies the LTI SDE in Eq.~\eqref{eq:gp_sde} with
\begin{equation}
    \mathbf{A} =
    \begin{bmatrix}
        \zero & \eye & \zero \\
        \zero & \zero & \eye \\
        \zero & \zero & \zero
    \end{bmatrix}_{18\times18}, \quad
    \mathbf{L} =
    \begin{bmatrix} \zero \\ \zero \\ \eye \end{bmatrix}_{18\times6},
\end{equation}
meaning the jerk $\dddot{\bm{\xi}}_i(t) = \mathbf{w}(t)$ is white noise. The local state transition matrix is given by
\begin{equation}
    \bm{\Phi}(t_{i+1}, t_i) =
    \begin{bmatrix}
        \eye & \Delta t\,\eye & \frac{1}{2}\Delta t^2\,\eye \\
        \zero & \eye & \Delta t\,\eye \\
        \zero & \zero & \eye
    \end{bmatrix}_{18\times18},
    \label{eq:Phi_classical}
\end{equation}
and the process noise covariance follows from the standard GP integral.

Although the conventional WNOJ formulation is not directly amenable to EKF-based estimation because of the coupling between translational and rotational states and its local covariance propagation, it provides the high-order kinematic variables required for highly dynamic LiDAR–inertial fusion, including pose, linear velocity and acceleration, and angular velocity and acceleration. In the following section, we reformulate the WNOJ prior by decoupling the translational and rotational dynamics and propagating the covariance in the global state space, thereby rendering it suitable as the prediction model of WNOJ-LIO.

\section{Decoupled WNOJ Process Model}
\label{sec:decoupled_wnoj}

In this section, we first derive the decoupled kinematic differential equations, then construct local pose variables and the corresponding LTI SDE. We finally map the local WNOJ recursion back to the global state and derive the covariance propagation needed by an EKF.

\subsection{Kinematic Decoupling via Coordinate Selection}
\label{sec:kinematic_decoupling}

The key insight is that the coupling between position and attitude derivatives depends on the choice of reference frame for the velocity variables. Using the Poisson equation, the attitude kinematics takes the form
\begin{equation}
    \dot{\mathbf{R}}_{wb} = \mathbf{R}_{wb}\,\skewop{\bm{\omega}^b_{wb}},
    \label{eq:poisson}
\end{equation}
which is independent of position. Expressing the velocity in a rotating body frame gives the position kinematics
\begin{equation}
    \dot{\mathbf{p}}^w_{wb} = -\mathbf{R}_{wb}\,\skewop{\mathbf{p}^b_{wb}}\,\bm{\omega}^b_{wb} + \mathbf{R}_{wb}\,\mathbf{v}^b_{wb},
\end{equation}
which depends on $\bm{\omega}^b_{wb}$, $\mathbf{R}_{wb}$, and $\mathbf{v}^b_{wb}$, creating a strong coupling. To decouple, we instead express velocity in the world frame and obtain the fully decoupled system
\begin{equation}
    \dot{\mathbf{p}}^w_{wb}(t) = \mathbf{v}^w_{wb}(t), \qquad
    \dot{\mathbf{R}}_{wb}(t) = \mathbf{R}_{wb}(t)\,\skewop{\bm{\omega}^b_{wb}(t)}.
    \label{eq:decoupled_diff}
\end{equation}
Introducing the acceleration variable $\mathbf{a}^w_{wb}$ and angular-acceleration variable $\bm{\alpha}^b_{wb}$ and imposing the WNOJ prior gives the continuous-time system
\begin{equation}
    \begin{aligned}
        \dot{\mathbf{p}}^w_{wb}(t) &= \mathbf{v}^w_{wb}(t), \\
        \dot{\mathbf{v}}^w_{wb}(t) &= \mathbf{a}^w_{wb}(t), \\
        \dot{\mathbf{a}}^w_{wb}(t) &= \mathbf{w}^w_{\mathbf{a}}(t), \quad
        \mathbf{w}^w_{\mathbf{a}} \sim \mathcal{GP}(\zero, \mathbf{q}_{\mathbf{a}}\,\delta(t-\tau)), \\
        \dot{\mathbf{R}}_{wb}(t) &= \mathbf{R}_{wb}(t)\,\skewop{\bm{\omega}^b_{wb}(t)}, \\
        \dot{\bm{\omega}}^b_{wb}(t) &= \bm{\alpha}^b_{wb}(t), \\
        \dot{\bm{\alpha}}^b_{wb}(t) &= \mathbf{w}^b_{\bm{\alpha}}(t), \quad
        \mathbf{w}^b_{\bm{\alpha}} \sim \mathcal{GP}(\zero, \mathbf{q}_{\bm{\alpha}}\,\delta(t-\tau)).
    \end{aligned}
    \label{eq:wnoj_continuous}
\end{equation}
Here, $\mathbf{p}^w_{wb}$, $\mathbf{v}^w_{wb}$, and $\mathbf{a}^w_{wb}$ are expressed in the world frame, while $\bm{\omega}^b_{wb}$ and $\bm{\alpha}^b_{wb}$ are expressed in the body frame. The translational jerk noise $\mathbf{w}^w_{\mathbf{a}}$ is therefore defined in the world frame, and the angular-acceleration noise $\mathbf{w}^b_{\bm{\alpha}}$ is defined in the body frame. For notational convenience, the frame superscripts and subscripts are omitted in the following derivations unless they are needed for clarity. The position and attitude subsystems are fully decoupled: position kinematics lives entirely in $\R^3$ and attitude kinematics on $\SO(3)$.

\subsection{Local Pose Variables and the Decoupled LTI SDE}

The decoupled global model above separates translation and rotation, but the EKF still requires a vector-space covariance recursion. We therefore introduce local pose variables over each propagation interval. This construction keeps the WNOJ prior linear in the tangent space while preserving the global-state prediction on $\R^3 \times \SO(3)$. Since $\mathbf{R} \in \SO(3)$ is not a vector-space quantity, we linearize in the local tangent space. For the time interval $[t_i, t_{i+1})$, we define the local pose variables as
\begin{equation}
    \delta\mathbf{p}_i(t) = \mathbf{p}(t) - \mathbf{p}(t_i), \qquad
    \delta\bm{\phi}_i(t) = \log\!\left(\mathbf{R}_i^{-1}\,\mathbf{R}(t)\right),
    \label{eq:local_vars}
\end{equation}
where $\mathbf{R}_i := \mathbf{R}(t_i)$. Differentiation yields the local velocities
\begin{equation}
    \dot{\delta\mathbf{p}}_i(t) = \mathbf{v}(t), \qquad
    \dot{\delta\bm{\phi}}_i(t) = \mathbf{J}_r\!\left(\delta\bm{\phi}_i(t)\right)^{-1}\bm{\omega}(t).
    \label{eq:local_velocity}
\end{equation}
Using the first-order approximation as Eq.~\eqref{eq:jr_approx}, the local angular acceleration is approximated as
\begin{equation}
    \ddot{\delta\bm{\phi}}_i(t) \approx
    \tfrac{1}{2}\skewop{\dot{\delta\bm{\phi}}_i(t)}\,\bm{\omega}(t)
    + \mathbf{J}_r^{-1}\!\left(\delta\bm{\phi}_i(t)\right)\bm{\alpha}(t).
    \label{eq:local_angular_acc}
\end{equation}

The augmented local state is partitioned as
\begin{equation}
    \bm{\gamma}_i(t) =
    \begin{bmatrix} \bm{\gamma}_{\mathbf{p},i}(t) \\ \bm{\gamma}_{\bm{\phi},i}(t) \end{bmatrix}, \quad
    \bm{\gamma}_{\mathbf{p},i} =
    \begin{bmatrix} \delta\mathbf{p}_i \\ \dot{\delta\mathbf{p}}_i \\ \ddot{\delta\mathbf{p}}_i \end{bmatrix}, \quad
    \bm{\gamma}_{\bm{\phi},i} =
    \begin{bmatrix} \delta\bm{\phi}_i \\ \dot{\delta\bm{\phi}}_i \\ \ddot{\delta\bm{\phi}}_i \end{bmatrix}.
    \label{eq:gamma_def}
\end{equation}
Under the decoupled WNOJ prior, the $\dddot{\delta\mathbf{p}}_i$ and $\dddot{\delta\bm{\phi}}_i$ are modeled as independent white-noise Gaussian processes with power spectral density $\mathbf{q}_{\mathbf{p}}$ and $\mathbf{q}_{\boldsymbol{\phi}}$, respectively.  Specifically,
\begin{equation}
    \mathbf{w}(t) =
    \begin{bmatrix}
        \mathbf{w}_{\mathbf{p}}(t) \\
        \mathbf{w}_{\boldsymbol{\phi}}(t)
    \end{bmatrix}, \qquad
    \begin{aligned}
        \mathbf{w}_{\mathbf{p}}(t) &\sim \mathcal{GP}\left( \mathbf{0}, \mathbf{q}_{\mathbf{p}}\delta(t-\tau)  \right), \\
        \mathbf{w}_{\boldsymbol{\phi}}(t) &\sim \mathcal{GP}\left( \mathbf{0}, \mathbf{q}_{\boldsymbol{\phi}}\delta(t-\tau) \right).
    \end{aligned}
\end{equation}

The resulting LTI SDE exhibits the block-diagonal structure
\begin{equation}
    \mathbf{A} =
    \begin{bmatrix}
        \mathbf{A}_{\mathbf{p}} & \zero_{9\times9} \\
        \zero_{9\times9} & \mathbf{A}_{\bm{\phi}}
    \end{bmatrix}, \quad
    \mathbf{L} =
    \begin{bmatrix}
        \mathbf{L}_{\mathbf{p}} & \zero_{9\times3} \\
        \zero_{9\times3} & \mathbf{L}_{\bm{\phi}}
    \end{bmatrix}, 
    \label{eq:decoupled_AL}
\end{equation}
where $\mathbf{A}_{\mathbf{p}} = \mathbf{A}_{\bm{\phi}}$ and $\mathbf{L}_{\mathbf{p}} = \mathbf{L}_{\bm{\phi}}$ have the same form as the classical WNOJ matrices given by
\begin{equation}
    \mathbf{A}_{\mathbf{p}} =
    \begin{bmatrix} \zero & \eye & \zero \\ \zero & \zero & \eye \\ \zero & \zero & \zero \end{bmatrix}_{9\times9}, \quad
    \mathbf{L}_{\mathbf{p}} =
    \begin{bmatrix} \zero \\ \zero \\ \eye \end{bmatrix}_{9\times3}.
\end{equation}

\begin{figure}[htbp]
	\centering
	\includegraphics[width=\columnwidth]{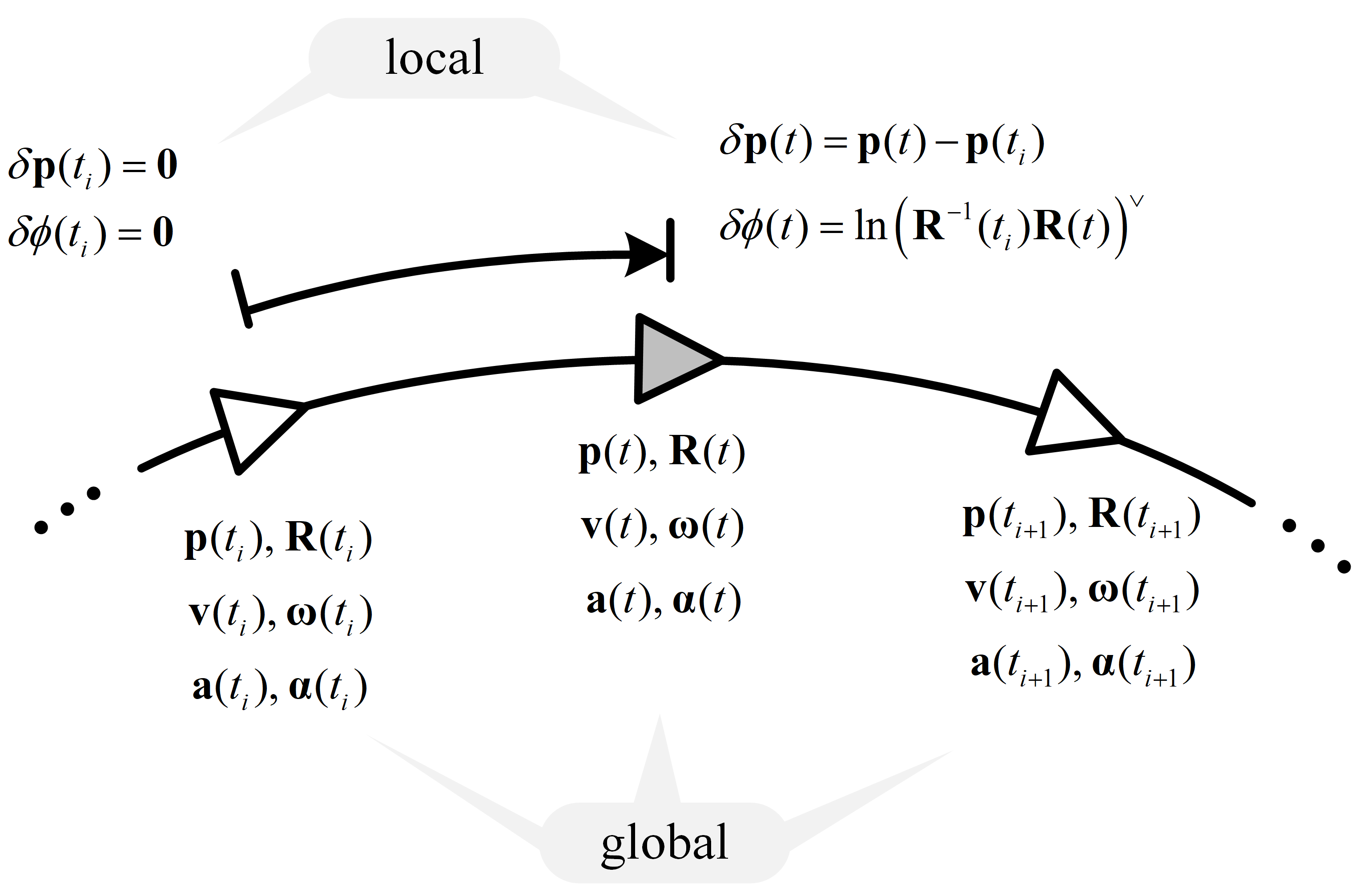}
	\caption{Global and local coordinate frames. The world frame $w$ serves as the global reference. The platform body frame $b$ (coinciding with the lidar frame) defines the pose to be estimated. Local pose variables are defined in the tangent space of $\SO(3)$ at the reference attitude.}
	\label{fig:global_local}
\end{figure}

\subsection{Discrete-Time System Equations} \label{sec:discrete_system_local_to_global}

The continuous-time local SDE is used over each finite propagation interval. We next discretize this local model over $[t_i,t_{i+1}]$ and relate the initialized local variables to the global state, so that the resulting mean and covariance can be used in the EKF prediction step.

The local state transition from $t_i$ to $t_{i+1}$ is expressed as
\begin{equation}
    \bm{\gamma}_i(t_{i+1}) = \bm{\Phi}(t_{i+1},t_i)\,\bm{\gamma}_i(t_i) + \mathbf{w}_{t_i},
    \label{eq:discrete_local}
\end{equation}
where the transition matrix takes the following block-diagonal form
\begin{equation}
    \bm{\Phi}(t_{i+1},t_i) =
    \begin{bmatrix}
        \bm{\Phi}_{\mathbf{p}} & \zero_{9\times9} \\
        \zero_{9\times9} & \bm{\Phi}_{\bm{\phi}}
    \end{bmatrix},
    \label{eq:Phi_decoupled}
\end{equation}
with $\bm{\Phi}_{\mathbf{p}} = \bm{\Phi}_{\bm{\phi}}$ having the same form as Eq.~\eqref{eq:Phi_classical} but at the $9\times9$ level. $\mathbf w_{t_i} = \int_{t_i}^{t_{i+1}} \bm\Phi(t_{i+1},\tau)\mathbf L\mathbf w(\tau)\,d\tau$.

At the initial time $t_i$, the local pose variables satisfy $\delta\mathbf{p}_i(t_i) = \zero$, $\delta\bm{\phi}_i(t_i) = \zero$, and $\mathbf{J}_r(\zero) = \eye$. It follows that
\begin{equation}
    \bm{\gamma}_i(t_i) =
    \begin{bmatrix}
        \zero \\ \mathbf{v}(t_i) \\ \mathbf{a}(t_i) \\ \hline
        \zero \\ \bm{\omega}(t_i) \\ \bm{\alpha}(t_i)
    \end{bmatrix},
    \quad
    \mathbf{P}_{\bm{\gamma}_i(t_i)} = \diag\!\left(\zero_3,\,\mathbf{P}_{\mathbf{v}},\,\mathbf{P}_{\mathbf{a}},\,\zero_3,\,\mathbf{P}_{\bm{\omega}},\,\mathbf{P}_{\bm{\alpha}}\right).
    \label{eq:gamma_initial}
\end{equation}


The local-state covariance propagates according to
\begin{equation}
    \mathbf{P}_{\bm{\gamma}_i(t_{i+1})} =
    \bm{\Phi}(t_{i+1},t_i)\,\mathbf{P}_{\bm{\gamma}_i(t_i)}\,\bm{\Phi}(t_{i+1},t_i)^\T + \mathbf{Q}_{t_i},
    \label{eq:cov_local}
\end{equation}
where $ \mathbf{Q}_{t_i} = \int_{t_i}^{t_{i+1}}  \bm{\Phi}(t_{i+1},\tau) \mathbf{L}\mathbf{Q}_c\mathbf{L}^{\T} \bm{\Phi}(t_{i+1},\tau)^{\T} \dd\tau$, $\mathbf{Q}_c = \diag \left( \mathbf{q}_{\mathbf{p}}, \mathbf{q}_{\bm{\phi}} \right)$. 
Because the translational and rotational jerk processes are independent, the discrete-time process-noise covariance takes the block-diagonal form
\begin{equation}
    \mathbf{Q}_{t_i} =
    \begin{bmatrix}
        \mathbf{Q}_{\mathbf{p},t_i} & \zero_{9\times9} \\
        \zero_{9\times9} & \mathbf{Q}_{\bm{\phi},t_i}
    \end{bmatrix},
    \label{eq:local_Q_blocks}
\end{equation}
where both covariance blocks share the coefficient matrix
\begin{equation}
    \mathbf{C}_{i} =
    \begin{bmatrix}
        \frac{\Delta t^5}{20} & \frac{\Delta t^4}{8} & \frac{\Delta t^3}{6} \\
        \frac{\Delta t^4}{8} & \frac{\Delta t^3}{3} & \frac{\Delta t^2}{2} \\
        \frac{\Delta t^3}{6} & \frac{\Delta t^2}{2} & \Delta t
    \end{bmatrix}.
    \label{eq:wnoj_cov_coeff}
\end{equation}
The translational and rotational components are then
\begin{equation}
    \mathbf{Q}_{\mathbf{p},t_i}
    = \mathbf{C}_{i}\otimes\mathbf{q}_{\mathbf{p}},
    \qquad
    \mathbf{Q}_{\bm{\phi},t_i}
    = \mathbf{C}_{i}\otimes\mathbf{q}_{\bm{\phi}},
    \label{eq:local_Q_components}
\end{equation}
where $\otimes$ denotes the Kronecker product.

Thus, translational and rotational process noise are injected through the local WNOJ acceleration and angular-acceleration channels. The global EKF covariance is obtained later by mapping this local covariance through the global-state Jacobian.
\subsection{Global WNOJ State and Manifold Structure}
\label{sec:wnoj_state}

The base WNOJ state at time $t_i$ is defined on the composite manifold as
\begin{equation}
    \mathbf{X}_i = \{\mathbf{p}_i,\, \mathbf{v}_i,\, \mathbf{a}_i,\, \mathbf{R}_i,\, \bm{\omega}_i,\, \bm{\alpha}_i\}
    \in \mathcal{M} = \R^9 \times \SO(3) \times \R^6,
    \label{eq:state_def}
\end{equation}
with $\dim(\mathcal{M}) = 18$. The $\boxplus$ operator applies a tangent-space perturbation according to
\begin{equation}
    \mathbf{X} \boxplus \delta\mathbf{X}
    = \begin{bmatrix} \mathbf{p} + \delta\mathbf{p} \\ \mathbf{v} + \delta\mathbf{v} \\ \mathbf{a} + \delta\mathbf{a} \\
    \mathbf{R}\,\exp(\delta\bm{\phi}) \\ \bm{\omega} + \delta\bm{\omega} \\ \bm{\alpha} + \delta\bm{\alpha} \end{bmatrix}.
    \label{eq:boxplus}
\end{equation}
The covariance is represented in the tangent space $\R^{18}$, with the attitude block parameterized in $\so(3)$ via $\bm{\phi} = \log(\mathbf{R})$.

\subsection{Global State Prediction with WNOJ Prior}
\label{sec:state_prediction}

The local WNOJ model in Subsection~\ref{sec:discrete_system_local_to_global} provides the relationship between the local state $\bm{\gamma}_i(t)=\left[\delta\mathbf{p}_i^\T,\dot{\delta\mathbf{p}}_i^\T,\ddot{\delta\mathbf{p}}_i^\T,\delta\bm{\phi}_i^\T,\dot{\delta\bm{\phi}}_i^\T,\ddot{\delta\bm{\phi}}_i^\T\right]^\T$ and the global state at $t_i$. Substituting Eq.~\eqref{eq:gamma_initial} into Eq.~\eqref{eq:discrete_local} yields the predicted local state
\begin{equation}
    \bm{\gamma}_i(t_{i+1}) =
    \begin{bmatrix}
        \mathbf{v}(t_i)\Delta t + \frac{1}{2}\mathbf{a}(t_i)\Delta t^2 \\
        \mathbf{v}(t_i) + \mathbf{a}(t_i)\Delta t \\
        \mathbf{a}(t_i) \\ \hline
        \bm{\omega}(t_i)\Delta t + \frac{1}{2}\bm{\alpha}(t_i)\Delta t^2 \\
        \bm{\omega}(t_i) + \bm{\alpha}(t_i)\Delta t \\
        \bm{\alpha}(t_i)
    \end{bmatrix}.
    \label{eq:gamma_predict}
\end{equation}

The global state prediction is obtained by combining Eq.~\eqref{eq:gamma_predict} with the relationship between local and global variables. For the position part, this gives
\begin{subequations}
\begin{align}
    \mathbf{p}(t_{i+1}) &= \mathbf{p}(t_i) + \mathbf{v}(t_i)\Delta t + \tfrac{1}{2}\mathbf{a}(t_i)\Delta t^2, \\
    \mathbf{v}(t_{i+1}) &= \mathbf{v}(t_i) + \mathbf{a}(t_i)\Delta t, \\
    \mathbf{a}(t_{i+1}) &= \mathbf{a}(t_i).
\end{align}
    \label{eq:pos_predict}
\end{subequations}

For the attitude part, the local attitude increment is given by
\begin{equation}
    \delta\bm{\phi}_i(t_{i+1}) = \bm{\omega}(t_i)\Delta t + \tfrac{1}{2}\bm{\alpha}(t_i)\Delta t^2,
    \label{eq:delta_phi_predict}
\end{equation}
and the rotation matrix is predicted as
\begin{equation}
    \mathbf{R}(t_{i+1}) = \mathbf{R}(t_i)\,\exp\!\left(\bm{\omega}(t_i)\Delta t + \tfrac{1}{2}\bm{\alpha}(t_i)\Delta t^2\right).
    \label{eq:R_predict}
\end{equation}

For the angular velocity, combining $\dot{\delta\bm{\phi}}_i(t_{i+1}) = \mathbf{J}_r^{-1}(\delta\bm{\phi}_i(t_{i+1}))\,\bm{\omega}(t_{i+1})$ with the first-order approximation gives
\begin{multline}
    \bm{\omega}(t_{i+1}) = \mathbf{J}_r\!\left(\delta\bm{\phi}_i(t_{i+1})\right)
    \left(\bm{\omega}(t_i) + \bm{\alpha}(t_i)\Delta t\right)
    \\
    \approx \bm{\omega}(t_i) + \bm{\alpha}(t_i)\Delta t
    - \tfrac{1}{2}\skewop{\bm{\omega}(t_i)}\bm{\alpha}(t_i)\Delta t^2 \\
    - \tfrac{1}{4}\skewop{\bm{\alpha}(t_i)}\bm{\omega}(t_i)\Delta t^2.
    \label{eq:omega_predict}
\end{multline}

For angular acceleration, the local relationship is $\ddot{\delta\bm{\phi}}_i(t_{i+1})=\bm{\alpha}(t_i)$ under the WNOJ prior, while $\ddot{\delta\bm{\phi}}_i(t_{i+1}) \approx \mathbf{J}_r^{-1}(\delta\bm{\phi}_i)\bm{\alpha}(t_{i+1})$. Using the first-order approximation of $\mathbf{J}_r$ gives the angular-acceleration prediction
\begin{equation}
    \bm{\alpha}(t_{i+1}) \approx \bm{\alpha}(t_i) - \tfrac{1}{2}\skewop{\bm{\omega}(t_i)}\bm{\alpha}(t_i)\Delta t.
    \label{eq:alpha_predict}
\end{equation}

Equations~\eqref{eq:pos_predict}-\eqref{eq:alpha_predict} constitute the WNOJ global-state process model. Note that the angular velocity and acceleration propagation include correction terms involving $\skewop{\bm{\omega}}\bm{\alpha}$ and $\skewop{\bm{\alpha}}\bm{\omega}$ that reflect the non-Euclidean geometry of $\SO(3)$, arising from the first-order approximation of the right Jacobian.

\subsection{Global Covariance Prediction}
\label{sec:cov_prediction}

The covariance prediction consists of two independent contributions. The first propagates the posterior state uncertainty at $t_i$ through the Jacobian of the deterministic global-state prediction. The second accounts for the uncertainty introduced by the WNOJ process noise over $[t_i,t_{i+1}]$. As derived in Subsection~\ref{sec:discrete_system_local_to_global}, $\mathbf{w}_{t_i}$ is the integrated process-noise increment of the local endpoint state $\bm{\gamma}_i(t_{i+1})$, with zero mean and covariance $\mathbf{Q}_{t_i}$.

The covariance of rotation in global state is represented in the Lie algebra space $\bm{\phi} = \log(\mathbf{R})$. Using the BCH linear approximation, $\bm{\phi}(t_{i+1}) = \bm{\phi}(t_i) + \mathbf{J}_r^{-1}(\bm{\phi}_i)\left(\bm{\omega}(t_i)\Delta t + \frac{1}{2}\bm{\alpha}(t_i)\Delta t^2\right)$. Let $\mathbf{x}$ denotes the global state vector $\mathbf{x}(t) = {[\mathbf{p}^\T, \mathbf{v}^\T, \mathbf{a}^\T, \bm{\phi}^\T, \bm{\omega}^\T, \bm{\alpha}^\T]}^\T \in \R^{18}$. Combining the deterministic state prediction with the local process-noise increment gives the linearized error propagation
\begin{equation}
    \mathbf{x}(t_{i+1}) = \bm{\Phi}_{i+1}'\,\mathbf{x}(t_i) + \bm{\Psi}_{i+1}'\,\mathbf{w}_{t_i},
    \label{eq:linearized_state}
\end{equation}
where the state transition Jacobian $\bm{\Phi}_{i+1}'$ is
\begin{equation}
    \resizebox{\columnwidth}{!}{$\displaystyle
    \bm{\Phi}_{i+1}' =
    \begin{bmatrix}
        \eye & \Delta t\,\eye & \frac{1}{2}\Delta t^2\,\eye & \zero & \zero & \zero \\
        \zero & \eye & \Delta t\,\eye & \zero & \zero & \zero \\
        \zero & \zero & \eye & \zero & \zero & \zero \\[4pt]
        \zero & \zero & \zero & \eye & \Delta t\,\mathbf{J}_r^{-1} & \frac{1}{2}\Delta t^2\,\mathbf{J}_r^{-1} \\
        \zero & \zero & \zero & \frac{1}{4}\Delta t^2\skewop{\bm{\alpha}} & \eye + \frac{1}{4}\Delta t^2\skewop{\bm{\alpha}} & \Delta t\,\eye - \frac{1}{4}\Delta t^2\skewop{\bm{\omega}} \\
        \zero & \zero & \zero & \zero & \frac{1}{2}\Delta t\,\skewop{\bm{\alpha}} & \eye - \frac{1}{2}\Delta t\,\skewop{\bm{\omega}}
    \end{bmatrix}$},
    \label{eq:Phi_prime}
\end{equation}
where all skew-symmetric terms are evaluated at $\bm{\omega}(t_i)$ and $\bm{\alpha}(t_i)$, and $\mathbf{J}_r^{-1} = \mathbf{J}_r^{-1}(\bm{\phi}_i)$. The matrix $\bm{\Psi}_{i+1}'$ directly maps the integrated local-state process noise into the global-state vector. It is defined as the local-to-global Jacobian and, under the same first-order approximation used in the state prediction, is given by
\begin{equation}
    \resizebox{\columnwidth}{!}{$\displaystyle
    \bm{\Psi}_{i+1}'
    \triangleq
    \frac{\partial\mathbf{x}(t_{i+1})}
    {\partial\bm{\gamma}_i(t_{i+1})^\T}
    \approx
    \begin{bmatrix}
        \eye & \zero & \zero & \zero & \zero & \zero \\
        \zero & \eye & \zero & \zero & \zero & \zero \\
        \zero & \zero & \eye & \zero & \zero & \zero \\[4pt]
        \zero & \zero & \zero & \mathbf{J}_r^{-1}(\bm{\phi}_i) & \zero & \zero \\
        \zero & \zero & \zero & \tfrac{1}{2}\skewop{\dot{\delta\bm{\phi}}_i(t_{i+1})} & \mathbf{J}_r\!\left(\delta\bm{\phi}_i(t_{i+1})\right) & \zero \\
        \zero & \zero & \zero & \tfrac{1}{2}\skewop{\ddot{\delta\bm{\phi}}_i(t_{i+1})} & \zero & \mathbf{J}_r\!\left(\delta\bm{\phi}_i(t_{i+1})\right)
    \end{bmatrix}$},
    \label{eq:Psi_prime}
\end{equation}

The EKF covariance prediction is then
\begin{equation}
    \check{\mathbf{P}}_{i+1} = \bm{\Phi}_{i+1}'\,\hat{\mathbf{P}}_i\,(\bm{\Phi}_{i+1}')^\T
    + \bm{\Psi}_{i+1}'\,\mathbf{Q}_{t_i}\,(\bm{\Psi}_{i+1}')^\T,
    \label{eq:cov_predict}
\end{equation}
where $\hat{\mathbf{P}}_i$ is the posterior covariance at $t_i$, and $\check{\mathbf{P}}_{i+1}$ is the predicted covariance at $t_{i+1}$. 
The matrices $\bm{\Phi}_{i+1}'$ and $\bm{\Psi}_{i+1}'$ are therefore the state transition and local-to-global process-noise mapping blocks used by the WNOJ-LIO EKF in the next section.

\section{WNOJ-LIO-Based EKF for LiDAR-IMU Fusion}
\label{sec:ekf_design}

The decoupled WNOJ model defines the base motion prior, but a LiDAR-IMU estimator must also include sensor-specific states, measurement models, and update scheduling. This section instantiates the WNOJ process model in an EKF. The resulting WNOJ-LIO estimator augments the 18-dimensional WNOJ state when a sensor requires additional variables, predicts the WNOJ and extension blocks with a partitioned process model, and updates the augmented state with high-rate IMU and low-rate LiDAR measurements.

\subsection{Augmented EKF State and Process Model}
\label{sec:augmented_ekf_state}

The WNOJ motion state in Eq.~\eqref{eq:state_def} is the base state shared by all sensor configurations. Additional variables required by a sensor model are collected in an extension vector $\mathbf{x}_{\mathrm{ext}}$. The augmented estimator state is defined as
\begin{equation}
    \mathbf{X}^{+}_i = \{\mathbf{X}^{\mathrm{WNOJ}}_i,\,\mathbf{x}_{\mathrm{ext},i}\}, \qquad
    \mathbf{X}^{\mathrm{WNOJ}}_i \equiv \mathbf{X}_i .
    \label{eq:aug_state_def}
\end{equation}
In the real-data WNOJ-LIO implementation, the IMU model requires the gravity vector. The nominal gravity is stored as a three-dimensional vector $\mathbf{g}\in\mathbb{R}^3$ with fixed magnitude, while its EKF error state is represented by a two-dimensional tangent perturbation $\delta\mathbf{g}\in\mathbb{R}^2$ on the $\mathbb{S}^2$ manifold. Thus, the extended state associated with gravity is given by $\mathbf{x}_{\mathrm{ext}} \triangleq \delta\mathbf{g}$. Other slowly varying calibration or sensor parameters can be appended in the same way, provided that their process and measurement Jacobian blocks are defined.

The augmented prediction is separated into the WNOJ process and sensor-extension blocks as
\begin{equation}
    \begin{aligned}
        \check{\mathbf{X}}^{\mathrm{WNOJ}}_{i+1}
        &= f_{\mathrm{WNOJ}}\!\left(\hat{\mathbf{X}}^{\mathrm{WNOJ}}_i,\Delta t\right), \\
        \check{\mathbf{x}}_{\mathrm{ext},i+1}
        &= f_{\mathrm{ext}}\!\left(\hat{\mathbf{x}}_{\mathrm{ext},i}\right).
    \end{aligned}
    \label{eq:aug_process}
\end{equation}
Here, $f_{\mathrm{WNOJ}}$ is given by Eqs.~\eqref{eq:pos_predict}-\eqref{eq:alpha_predict}. For the gravity extension used here, the nominal gravity direction is modeled as constant and the tangent-space error transition is identity.

The corresponding tangent-space transition matrix is partitioned as
\begin{equation}
    \mathbf{F}^{+}_{i+1} =
    \begin{bmatrix}
        \mathbf{F}_{\mathrm{WNOJ},i+1} & \zero \\
        \zero & \mathbf{F}_{\mathrm{ext},i+1}
    \end{bmatrix}, \qquad
    \mathbf{F}_{\mathrm{WNOJ},i+1}=\bm{\Phi}'_{i+1}.
    \label{eq:aug_F}
\end{equation}
For the gravity extension, $\mathbf{F}_{\mathrm{ext},i+1}=\eye_2$. The augmented covariance prediction is
\begin{equation}
    \check{\mathbf{P}}^{+}_{i+1}
    = \mathbf{F}^{+}_{i+1}\hat{\mathbf{P}}^{+}_{i}(\mathbf{F}^{+}_{i+1})^\T
    + \mathbf{Q}^{+}_{i},
    \label{eq:aug_cov_predict}
\end{equation}
with
\begin{equation}
    \mathbf{Q}^{+}_{i} =
    \diag\!\left(
    \bm{\Psi}'_{i+1}\mathbf{Q}_{t_i}(\bm{\Psi}'_{i+1})^\T,\,
    \mathbf{Q}_{\mathrm{ext},i}
    \right).
    \label{eq:aug_Q}
\end{equation}
Thus, the WNOJ block retains the covariance design derived in Section~\ref{sec:decoupled_wnoj}, while the extension block is added only when required by the sensor configuration. If a future extension state is coupled to the WNOJ motion in prediction, the off-diagonal blocks in Eq.~\eqref{eq:aug_F} can be filled by the corresponding process Jacobians.

\subsection{IMU Measurement Model (High-Frequency Update)}
\label{sec:measurement_imu}

Unlike conventional LiDAR-inertial methods that integrate IMU in the prediction step, we treat IMU as a \emph{measurement update} in the EKF. This enables the WNOJ prior to act as a high-order smoother for the raw IMU measurements, producing denoised estimates of acceleration and angular velocity.

An IMU measures the specific force $\mathbf{a}_m$ and angular velocity $\bm{\omega}_m$ in the body frame according to
\begin{equation}
    \mathbf{y}_{\text{IMU}} =
    \begin{bmatrix} \mathbf{a}_m \\ \bm{\omega}_m \end{bmatrix}
    =
    \begin{bmatrix} \mathbf{R}^\T(\mathbf{a} - \mathbf{g}) \\ \bm{\omega} \end{bmatrix}
    + \mathbf{n}_{\text{IMU}},
    \label{eq:measurement_imu}
\end{equation}
where $\mathbf{g}$ is the gravity vector in the world frame, estimated as an extended state in the real-data implementation, and $\mathbf{n}_{\text{IMU}} \sim \mathcal{N}(\zero, \bm{\Sigma}_{\text{IMU}})$ is the measurement noise. The measurement function is $\mathbf{h}_{\text{IMU}}(\mathbf{X}^{+}) = [\mathbf{R}^\T(\mathbf{a} - \mathbf{g});\; \bm{\omega}]$.

We do not append accelerometer and gyroscope biases to the state in this paper. In the racing-car data, the effective bias is entangled with temperature variation and vibration-induced components, and preliminary simulations showed that directly estimating such biases in this WNOJ update formulation was unreliable. Because WNOJ-LIO directly estimates acceleration and angular velocity and constrains them through IMU measurements, residual IMU bias is handled through the measurement residual and covariance.

The IMU Jacobian is partitioned according to the augmented state, $\mathbf{H}_{\text{IMU}}=[\mathbf{H}_{\text{IMU,WNOJ}}\;\mathbf{H}_{\text{IMU,ext}}]$. The WNOJ-state block contains non-zero columns for acceleration $\mathbf{a}$, rotation $\mathbf{R}$ (via $\bm{\phi}$), and angular velocity $\bm{\omega}$. For the accelerometer part, the Jacobian with respect to $\bm{\phi}$ involves $\skewop{\mathbf{R}^\T(\mathbf{a}-\mathbf{g})}\,\mathbf{J}_r(\bm{\phi})$ through the chain rule on $\SO(3)$, and with respect to $\mathbf{a}$ is simply $\mathbf{R}^\T$. For the gyroscope part, the Jacobian with respect to $\bm{\omega}$ is $\eye$, and all other WNOJ-state blocks are zero, giving
\begin{equation}
    \resizebox{\columnwidth}{!}{$\displaystyle
    \mathbf{H}_{\text{IMU,WNOJ}} =
    \begin{bmatrix}
        \zero_{3} & \zero_{3} & \mathbf{R}^\T & \skewop{\mathbf{R}^\T(\mathbf{a}-\mathbf{g})}\,\mathbf{J}_r & \zero_{3} & \zero_{3} \\
        \zero_{3} & \zero_{3} & \zero_{3} & \zero_{3} & \eye_{3} & \zero_{3}
    \end{bmatrix}$}.
    \label{eq:H_imu}
\end{equation}
If gravity is included in $\mathbf{x}_{\mathrm{ext}}$, let $\mathbf{B}_{g}\in\mathbb{R}^{3\times2}$ denote the orthonormal tangent basis at the current gravity vector on $\mathbb{S}^2$. The implementation uses the local retraction $\mathbf{g}\boxplus\delta\mathbf{g}=\exp(\mathbf{B}_g\delta\mathbf{g})\mathbf{g}$, whose first-order Jacobian is $\partial\mathbf{g}/\partial\delta\mathbf{g}=-\skewop{\mathbf{g}}\mathbf{B}_{g}$. Therefore, the IMU extension block is
\begin{equation}
    \mathbf{H}_{\text{IMU,ext}} =
    \begin{bmatrix}
        \mathbf{R}^\T\skewop{\mathbf{g}}\mathbf{B}_{g} \\
        \zero_{3\times2}
    \end{bmatrix}.
    \label{eq:H_imu_ext}
\end{equation}
This follows from $\partial\mathbf{R}^\T(\mathbf{a}-\mathbf{g})/\partial\mathbf{g}=-\mathbf{R}^\T$ and the $\mathbb{S}^2$ tangent-space gravity parametrization used in VINS-Mono~\cite{qinVINSMono2018}. The detailed $\mathbb{S}^2$ basis construction is omitted because this gravity parametrization is an implementation detail rather than a contribution of this paper.

The IMU measurement update executes at 200~Hz (each IMU sample triggers a prediction + update cycle). After the EKF update, the posterior state estimates $\hat{\mathbf{a}}$ and $\hat{\bm{\omega}}$ are the \emph{denoised} acceleration and angular velocity. This is fundamentally different from FAST-LIO-style approaches: rather than using raw $\mathbf{a}_m$ and $\bm{\omega}_m$ for forward propagation, the WNOJ-LIO refines these estimates through the high-order motion prior, filtering out high-frequency noise while preserving the true dynamic signal.

\subsection{Point Cloud Motion Compensation Using Posterior State History}
\label{sec:de_distortion}

During a LiDAR scan interval $[t_{\ell,s},t_{\ell,e}]$, each point $\mathbf{p}^{\mathrm{raw}}_{\ell,m}$ is acquired at its own timestamp $\tau_{\ell,m}\in[t_{\ell,s},t_{\ell,e}]$ and therefore corresponds to a different sensor pose. To compensate for the resulting motion distortion, WNOJ-LIO utilizes a history of posterior poses generated by IMU-rate EKF updates, rather than relying on poses obtained through direct integration of raw IMU measurements.

Let the stored IMU-update times be $\mathcal{T}_I=\{t_{I,i},t_{I,i+1},\ldots,t_{I,i+K}\}$. After each IMU update, the EKF stores the corresponding posterior pose
\begin{equation}
    \hat{\mathbf{T}}_{I,j} =
    \begin{bmatrix} \hat{\mathbf{R}}_{I,j} & \hat{\mathbf{p}}_{I,j} \\ \zero & 1 \end{bmatrix},
    \qquad j=i,\ldots,i+K,
    \label{eq:posterior_pose_history}
\end{equation}
where $\hat{\mathbf{R}}_{I,j}$ and $\hat{\mathbf{p}}_{I,j}$ denote the IMU-rate posterior orientation and position estimates, respectively. The arrival time of each LiDAR scan is taken as the reference time $t_{\ell,\mathrm{ref}}$ for motion compensation, and the pose predicted at this time is denoted by $\check{\mathbf{T}}_{\ell,\mathrm{ref}}$.

Since the adjacent posterior poses are separated by only a short time interval, a computationally efficient first-order interpolation scheme is adopted to compute the sensor pose at the LiDAR point acquisition time. For each LiDAR point timestamp $\tau_{\ell,m}$, two neighboring IMU posterior poses $\hat{\mathbf{T}}_{I,j}$ and $\hat{\mathbf{T}}_{I,j+1}$ are selected from the stored history. The interpolation ratio is
\begin{equation}
    \beta_{\ell,m} = \frac{\tau_{\ell,m}-t_{I,j}}{t_{I,j+1}-t_{I,j}}.
    \label{eq:interp_beta}
\end{equation}
The LiDAR point-time pose is reconstructed through linear interpolation of translation and geodesic interpolation of rotation
\begin{equation}
    \begin{aligned}
    \hat{\mathbf{p}}_{\ell,m} &= (1-\beta_{\ell,m})\hat{\mathbf{p}}_{I,j}
        + \beta_{\ell,m}\hat{\mathbf{p}}_{I,j+1}, \\
    \hat{\mathbf{R}}_{\ell,m} &= \hat{\mathbf{R}}_{I,j}
        \exp\!\left(\beta_{\ell,m} \log(\hat{\mathbf{R}}_{I,j}^\T\hat{\mathbf{R}}_{I,j+1})\right), \\
    \hat{\mathbf{T}}_{\ell,m} &=
    \begin{bmatrix} \hat{\mathbf{R}}_{\ell,m} & \hat{\mathbf{p}}_{\ell,m} \\ \zero & 1 \end{bmatrix}.
    \end{aligned}
    \label{eq:T_interpolate}
\end{equation}
Although the interpolation is performed locally between two adjacent posterior poses, the interpolated trajectory inherits the temporal smoothness imposed by the WNOJ prior as well as the corrections introduced by the IMU measurement updates.

Each raw point is subsequently transformed to the LiDAR reference frame associated with $\hat{\mathbf{T}}_{\ell,\mathrm{ref}}$
\begin{equation}
    \mathbf{p}^{\text{corr}}_{\ell,m} = \check{\mathbf{T}}_{\ell,\mathrm{ref}}^{-1}\,\hat{\mathbf{T}}_{\ell,m}\,\mathbf{p}^{\text{raw}}_{\ell,m}.
    \label{eq:de_distortion}
\end{equation}
Unlike FAST-LIO, which performs motion compensation using poses obtained through direct integration of raw IMU measurements, the proposed approach relies on a sequence of posterior poses generated by repeated IMU update steps. Consequently, high-frequency IMU noise is attenuated before entering the motion-compensation process, reducing the propagation of IMU-induced errors to the LiDAR measurements used in the subsequent point-to-plane EKF update.

\subsection{LiDAR Measurement Model (Low-Frequency Update)}
\label{sec:measurement_lidar}

After de-distortion, the LiDAR scan is aligned with the local map through point-to-plane registration, with map maintenance and local plane fitting performed following the FAST-LIO mapping strategy.
For each selected de-distorted point $m\in\mathcal{S}_{\ell}$, the implementation transforms $\mathbf{p}^{\text{corr}}_{\ell,m}$ to the world frame and fits a local plane from its nearest map points. Let this plane be $\mathbf{n}_{m}^{\T}\mathbf{x}+d_m=0$, where $\mathbf{n}_{m}$ is the world-frame unit normal and $d_m$ is the plane offset. The LiDAR measurement function and residual are
\begin{equation}
    \begin{aligned}
    h_{\text{LiDAR},m}(\mathbf{X}^{+}) &=
    \mathbf{n}_{m}^{\T}\!\left(\mathbf{R}\mathbf{p}^{\text{corr}}_{\ell,m}+\mathbf{p}\right)+d_m, \qquad m\in\mathcal{S}_{\ell},
    \end{aligned}
    \label{eq:measurement_lidar}
\end{equation}
where the zero point-to-plane distance is treated as the LiDAR measurement. The corresponding WNOJ-state measurement-function Jacobian used in the EKF update is
\begin{equation}
    \resizebox{\columnwidth}{!}{$\displaystyle
    \mathbf{H}_{\text{LiDAR,WNOJ},m} = \left[ \mathbf{n}_{m}^{\T},\; \zero^\T,\; \zero^\T,\;
    -\mathbf{n}_{m}^{\T}\mathbf{R}\skewop{\mathbf{p}^{\text{corr}}_{\ell,m}}\,\mathbf{J}_r(\bm{\phi}),\; \zero^\T,\; \zero^\T \right]$}.
    \label{eq:H_lidar}
\end{equation}
For the gravity-augmented WNOJ-LIO state, the LiDAR residual does not directly depend on the gravity extension, so $\mathbf{H}_{\text{LiDAR,ext}}=\zero$.

The LiDAR measurement update executes at 20~Hz (after each scan completion), correcting the position $\mathbf{p}$ and attitude $\mathbf{R}$ components of the state. Combined with the high-frequency IMU update, this dual-frequency architecture provides continuous state refinement.

\subsection{Dual-Frequency EKF Update}
\label{sec:dual_frequency}

The complete measurement update follows the standard EKF formulation on the augmented state. Given a measurement vector $\mathbf{y}_{i+1}$ with measurement function $\mathbf{h}(\mathbf{X}^{+}_{i+1})$ and noise covariance $\bm{\Sigma}_{\mathbf{y}}$, the residual and Jacobian are defined as
\begin{equation}
    \begin{aligned}
        \mathbf{r}_{i+1} &=
        \mathbf{y}_{i+1} - \mathbf{h}(\check{\mathbf{X}}^{+}_{i+1}), \\
        \mathbf{H}_{i+1} &=
        \begin{bmatrix}
            \mathbf{H}_{\mathrm{WNOJ},i+1} &
            \mathbf{H}_{\mathrm{ext},i+1}
        \end{bmatrix}.
    \end{aligned}
    \label{eq:H_partition}
\end{equation}
The Kalman update is
\begin{equation}
    \begin{aligned}
        \mathbf{K}_{i+1} &= \check{\mathbf{P}}^{+}_{i+1}\mathbf{H}_{i+1}^\T
        \left(\mathbf{H}_{i+1}\check{\mathbf{P}}^{+}_{i+1}\mathbf{H}_{i+1}^\T + \bm{\Sigma}_{\mathbf{y}}\right)^{-1}, \\
        \hat{\mathbf{P}}^{+}_{i+1} &= \left(\eye - \mathbf{K}_{i+1}\mathbf{H}_{i+1}\right)\check{\mathbf{P}}^{+}_{i+1}, \\
        \hat{\mathbf{X}}^{+}_{i+1} &= \check{\mathbf{X}}^{+}_{i+1} \boxplus^{+} \mathbf{K}_{i+1}\mathbf{r}_{i+1},
    \end{aligned}
    \label{eq:ekf_update}
\end{equation}
where $\boxplus^{+}$ applies the manifold perturbation to the WNOJ block and, for the gravity extension, applies the $\mathbb{S}^2$ retraction to the two-dimensional tangent perturbation. For the IMU update, $\mathbf{H}_{\mathrm{WNOJ}}=\mathbf{H}_{\text{IMU,WNOJ}}$ and $\mathbf{H}_{\mathrm{ext}}=\mathbf{H}_{\text{IMU,ext}}$ (Eqs.~\eqref{eq:H_imu}-\eqref{eq:H_imu_ext}). For the LiDAR update, $\mathbf{H}_{\mathrm{WNOJ}}=\mathbf{H}_{\text{LiDAR,WNOJ}}$ and $\mathbf{H}_{\mathrm{ext}}=\zero$ (Eq.~\eqref{eq:H_lidar}).

\section{Simulation Study}
\label{sec:simulation}

This section evaluates WNOJ-LIO through a controlled simulation against a FAST-LIO-style baseline inspired by FAST-LIO~\cite{xuFASTLIO2021, xuFASTLIO22022}. The baseline preserves the key FAST-LIO mechanism in which raw IMU measurements drive state prediction, while using the same simulated point-to-plane LiDAR measurements as WNOJ-LIO. This controlled setting removes mapping-related effects and isolates the influence of the inertial propagation strategy. The evaluation is organized around two inertial sensing conditions and one ablation: normal IMU noise, high IMU noise, and a WNOJ-LIO variant without IMU updates. This structure tests whether the proposed WNOJ prediction plus IMU-measurement-update design remains effective when raw inertial propagation becomes noisy, and separates the contribution of the WNOJ prior from that of the IMU measurements.

\subsection{Simulation Setup}

\begin{figure}[htbp]
    \centering
    \includegraphics[width=\columnwidth]{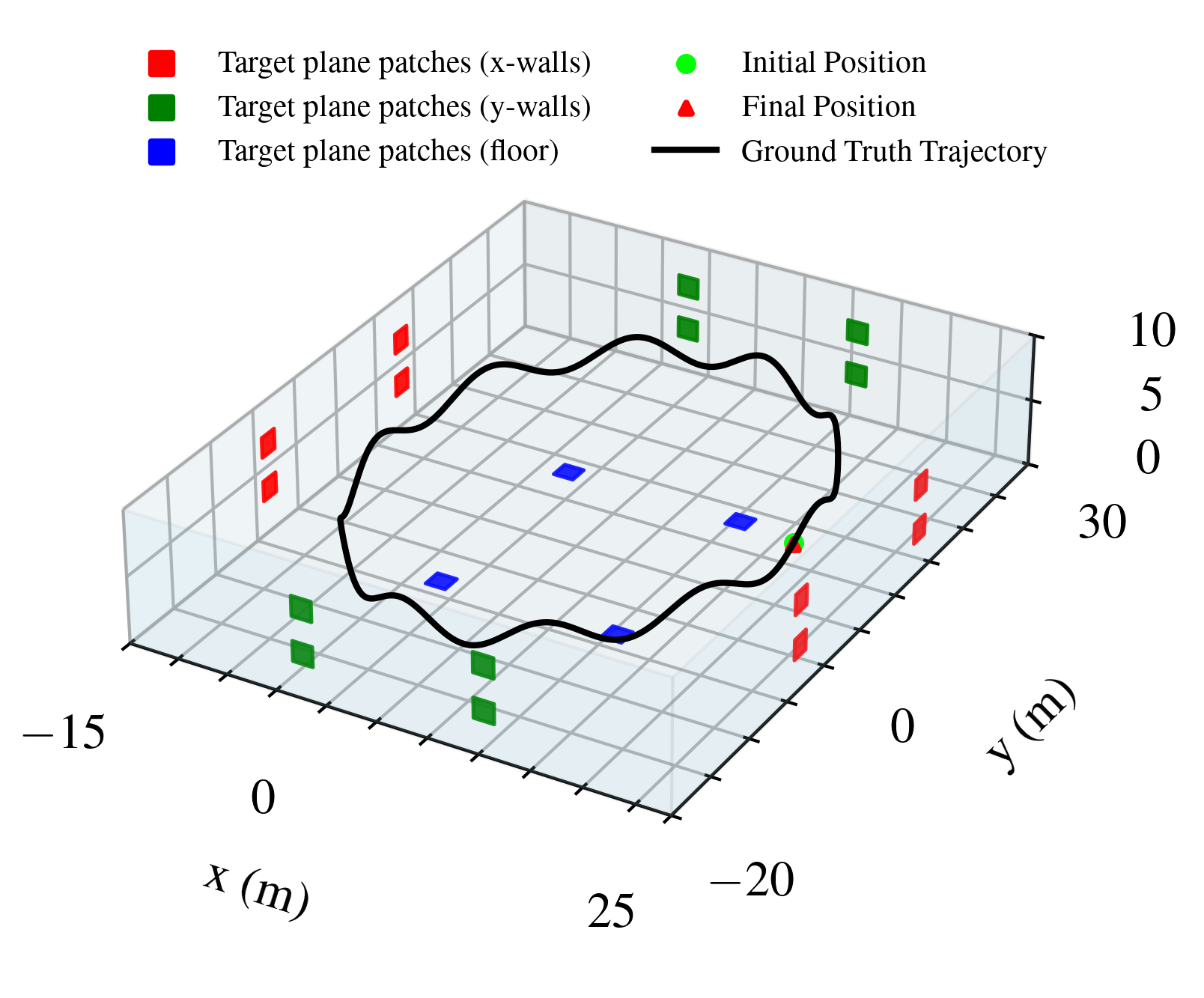}
    \caption{A $40\times50\times10$~m simulation environment. The black curve is the ground-truth trajectory. The small rectangles are target plane patches (20 in total) whose midpoints serve as LiDAR point-to-plane measurement sources, decoupling odometry estimation from mapping; patches are colored by location (red: $x$-walls, green: $y$-walls, blue: floor). Green and red markers denote the initial and final positions, respectively.}
    \label{fig:sim_setup}
\end{figure}

As shown in Fig.~\ref{fig:sim_setup}, a simulated environment of size $40\times50\times10$~m is constructed. To evaluate the influence of different IMU integration strategies on odometry estimation independently of mapping quality, 20 known planar patches are distributed on the surrounding walls and floor instead of using a dynamically built map. The centroids of these patches are used to generate point-to-plane LiDAR measurements, thereby decoupling odometry estimation from mapping errors.

A continuous-time six-degree-of-freedom ground-truth trajectory and the corresponding 200~Hz IMU measurements are generated using a public simulation framework\footnote{\url{https://github.com/robosu12/imu_data_simulation.git}}. To assess robustness across inertial sensing conditions, two IMU measurement sets are synthesized from the same continuous-time acceleration and angular-velocity ground truth. The normal-noise setting uses zero-mean white noise with standard deviations $\sigma_a = 0.0294$~m/s$^2$ and $\sigma_{\omega} = 0.00175$~rad/s, with accelerometer and gyroscope bias random-walk coefficients of $5\times10^{-4}$~(m/s$^2$)/$\sqrt{\text{s}}$ and $5\times10^{-5}$~(rad/s)/$\sqrt{\text{s}}$, respectively. The high-noise setting scales the IMU noise level to represent vibration-amplified MEMS measurements on high-dynamic platforms, where sensor output noise can increase substantially with mechanical excitation~\cite{xuAdaptive2023,hoangNoise2021,chenRobust2024}.

LiDAR measurements are simulated at 20~Hz, yielding a scan duration of 50~ms. During each scan, the 20 planar patch centroids are observed sequentially at uniformly spaced timestamps within the scan interval, yielding one LiDAR point per patch. The coordinates of these points are generated by evaluating the continuous-time trajectory at their respective acquisition times. Consequently, the resulting point cloud naturally exhibits intra-scan motion distortion. Independent Gaussian noise is further added to each LiDAR point measurement.

All methods use the same trajectory, planar patches, LiDAR timestamps, and noise realizations within each noise setting.

\subsection{Acceleration and Angular-Velocity Denoising Evaluation}

We first examine the effect of IMU noise on the estimated vehicle acceleration and angular velocity, which are used later for point cloud de-distortion. In this simulation, ground-truth vehicle acceleration and angular velocity are available, so the evaluation is not a direct comparison of IMU sensor outputs. In the FAST-LIO-style baseline, raw IMU acceleration and angular velocity are directly forward-integrated for state prediction, so IMU sensor noise propagates into the predicted trajectory and is subsequently transferred to LiDAR motion compensation. In contrast, WNOJ-LIO uses the IMU as a measurement update: the third-order WNOJ prior provides a smooth state prediction, and the noisy IMU measurement is fused through the Kalman gain to produce denoised posterior acceleration and angular-velocity estimates.

\begin{figure}[htbp]
    \centering
    \subfloat[]{
        \includegraphics[width=0.48\columnwidth]{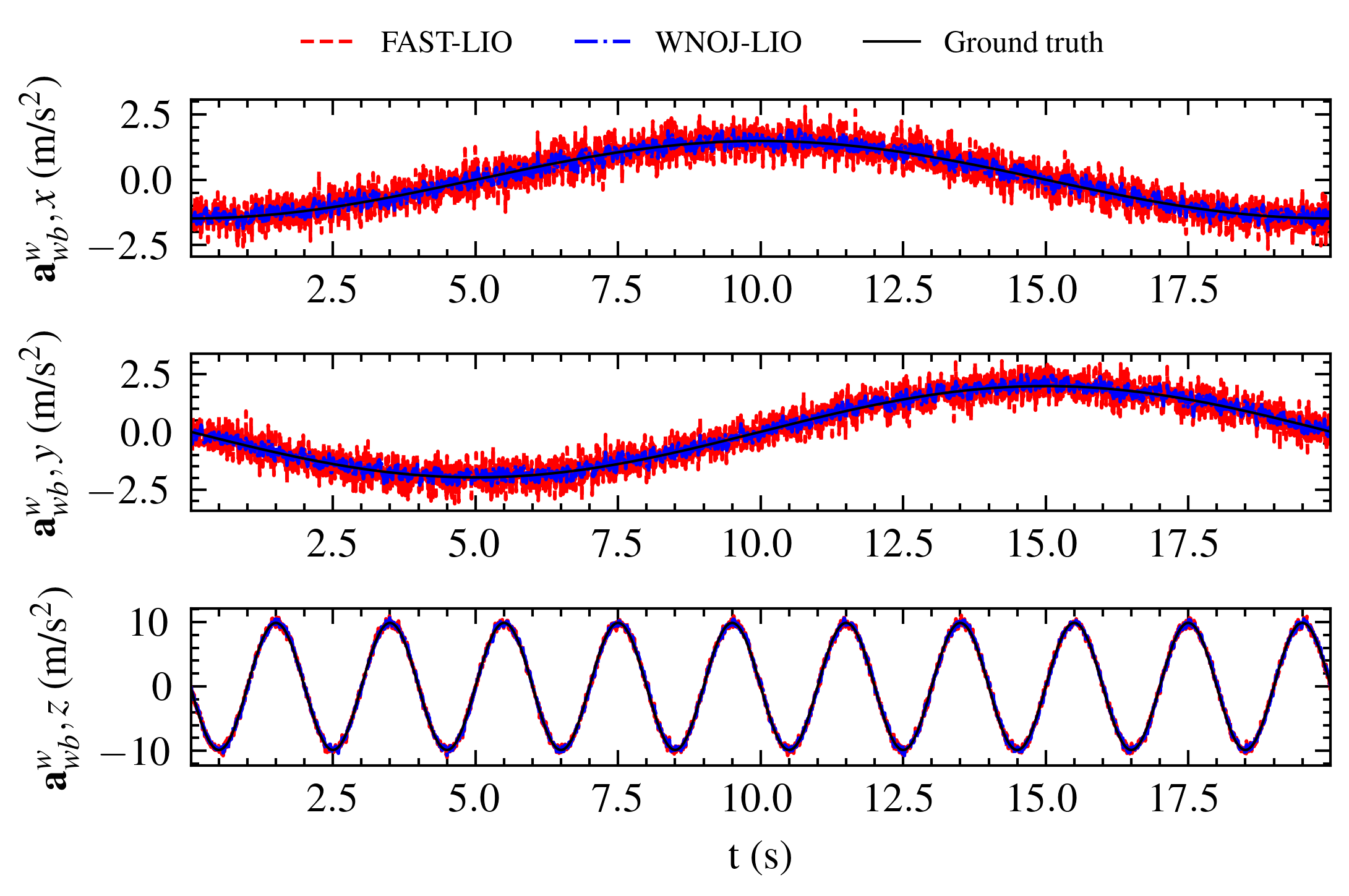}
    }
    \subfloat[]{
        \includegraphics[width=0.48\columnwidth]{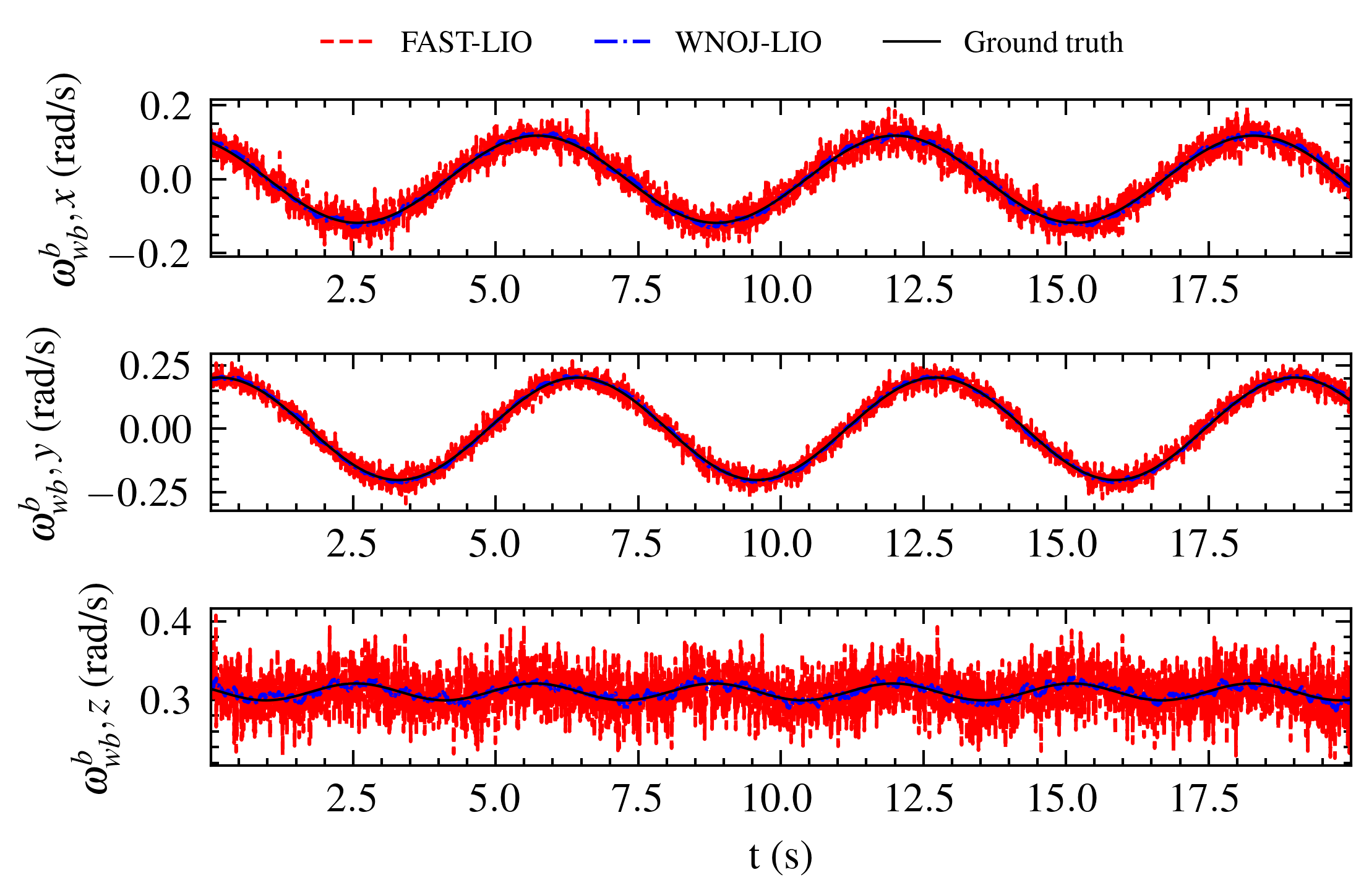}
    }
    \caption{Acceleration and angular-velocity estimation under normal IMU noise. (a) Acceleration. (b) Angular velocity. The FAST-LIO-style baseline retains noise from direct IMU propagation, while WNOJ-LIO estimates smoother acceleration and angular velocity through high-rate IMU measurement updates.}
    \label{fig:sim_imu_denoising_normal}
\end{figure}

\begin{figure}[htbp]
    \centering
    \subfloat[]{
        \includegraphics[width=0.48\columnwidth]{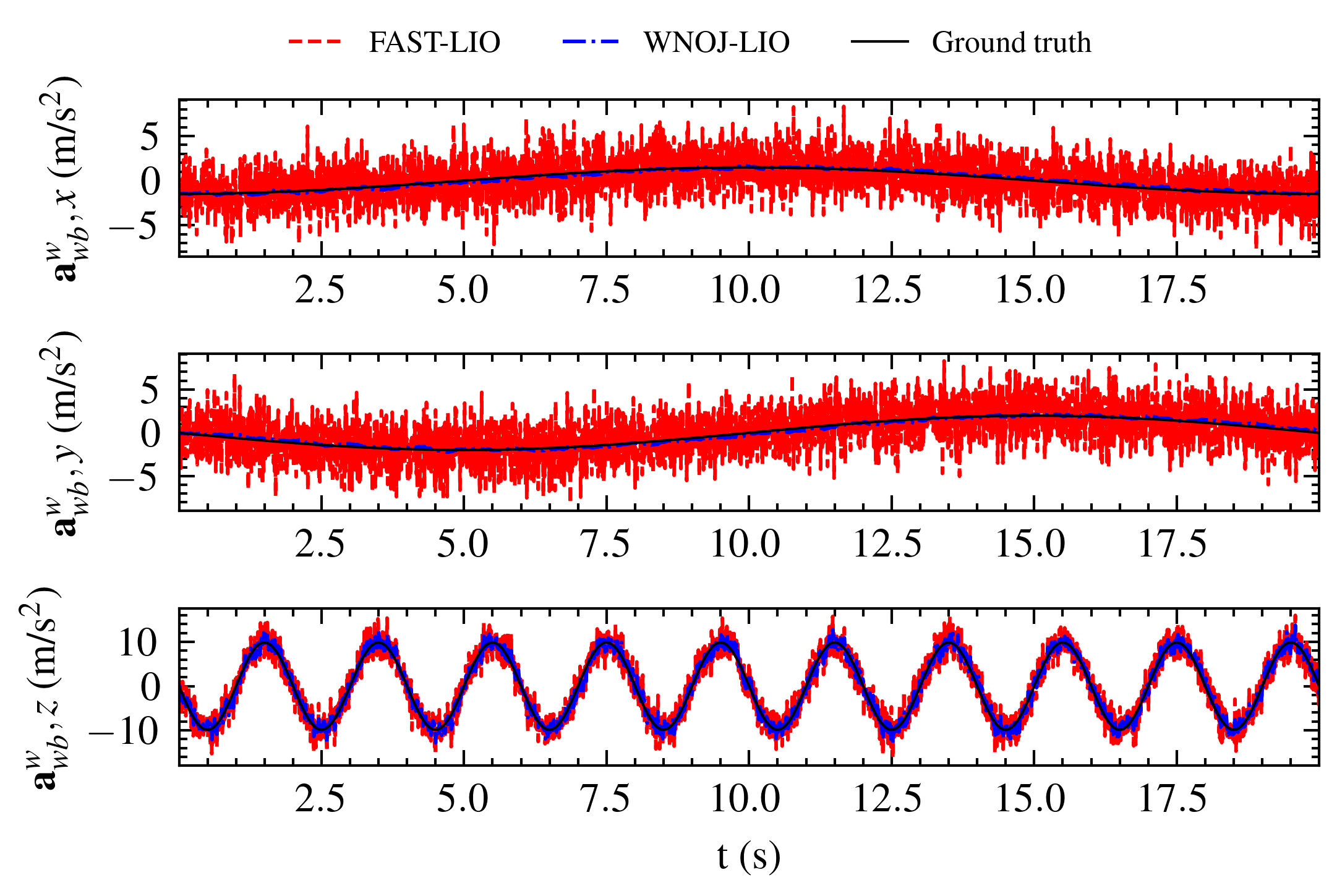}
    }
    \subfloat[]{
        \includegraphics[width=0.48\columnwidth]{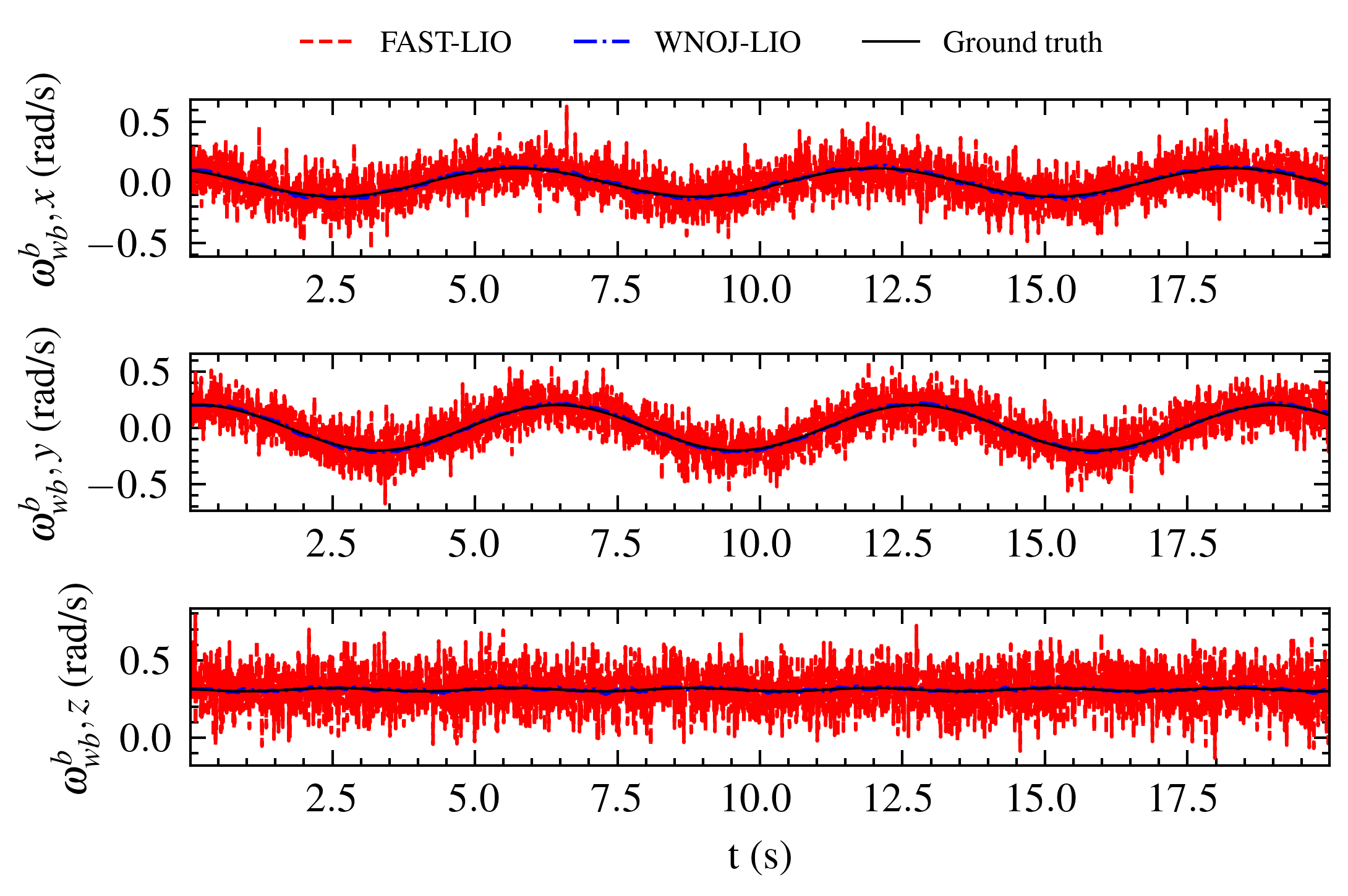}
    }
    \caption{Acceleration and angular-velocity estimation under large IMU noise. (a) Acceleration. (b) Angular velocity. Compared with the FAST-LIO-style baseline, WNOJ-LIO suppresses the amplified IMU-driven noise more effectively and preserves acceleration and angular-velocity estimates closer to the ground truth.}
    \label{fig:sim_imu_denoising_large}
\end{figure}

Figs.~\ref{fig:sim_imu_denoising_normal} and~\ref{fig:sim_imu_denoising_large} compare the estimated vehicle acceleration and angular velocity under normal and large IMU noise, respectively. In both settings, the FAST-LIO-style prediction closely tracks the noisy IMU-driven propagation, whereas the WNOJ-LIO posterior produces smoother acceleration and angular-velocity estimates closer to the ground truth. The difference becomes more visible under large IMU noise, where direct IMU propagation transfers high-frequency noise into the estimated acceleration and angular velocity.

Table~\ref{tab:imu_denoising} quantifies the acceleration and angular-velocity RMSE under both IMU noise settings. Under normal IMU noise, WNOJ-LIO reduces acceleration RMSE by $43.1\%$ and angular-velocity RMSE by $79.8\%$ compared with the FAST-LIO-style propagation. Under high IMU noise, the reductions increase to $69.1\%$ and $93.3\%$, respectively. These results show that treating the IMU as an EKF measurement is particularly beneficial when raw inertial propagation becomes noisy.

\begin{table}[htbp]
\caption{Acceleration and Angular-Velocity RMSE Under IMU Noise Settings}
\label{tab:imu_denoising}
\centering\footnotesize
\setlength{\tabcolsep}{4pt}
\begin{tabular}{lllcc}
\toprule
\multirow{2}{*}{Noise} & \multirow{2}{*}{Method} & \multirow{2}{*}{IMU role} & Acceleration & Angular velocity \\
 & & & (m/s$^2$) & (rad/s) \\
\midrule
\multirow{2}{*}{Normal} & FAST-LIO-style & Prediction & 0.7317 & 0.0431 \\
 & WNOJ-LIO & Update & \textbf{0.4164} & \textbf{0.0087} \\
\midrule
\multirow{2}{*}{High} & FAST-LIO-style & Prediction & 3.6178 & 0.2153 \\
 & WNOJ-LIO & Update & \textbf{1.1181} & \textbf{0.0143} \\
\bottomrule
\end{tabular}
\end{table}

\subsection{Point Cloud De-distortion Evaluation}

The denoised high-rate posterior state history directly affects LiDAR point cloud de-distortion quality. During each 50~ms scan period, both methods transform raw body-frame LiDAR points to the scan-end body frame using their respective state estimates. The FAST-LIO-style baseline uses forward-integrated IMU states, while WNOJ-LIO uses the high-frequency posterior pose history obtained after WNOJ prediction and IMU updates.

Table~\ref{tab:dewarp_rmse} reports the point-cloud RMSE after expressing each scan in the scan-end body frame. Without de-distortion, the raw point-cloud error is $0.2653$~m. Both methods reduce this error, but WNOJ-LIO remains lower under both noise settings, especially under high IMU noise.


\begin{table}[htbp]
\caption{Point-Cloud De-Distortion RMSE Under IMU Noise Settings}
\label{tab:dewarp_rmse}
\centering\footnotesize
\setlength{\tabcolsep}{4pt}
\begin{tabular}{lccc}
\toprule
\multirow{2}{*}{Noise} & Raw & FAST-LIO-style & WNOJ-LIO \\
 & No de-distortion & de-distorted & de-distorted \\
\midrule
Normal & 0.2653 & 0.0651 & \textbf{0.0555} \\
High & 0.2653 & 0.0964 & \textbf{0.0598} \\
\bottomrule
\end{tabular}
\begin{flushleft}\vspace{-6pt}
\footnotesize RMSE is computed between each corrected point and the noiseless ground-truth point expressed in the scan-end body frame. The raw no-de-distortion value is identical for the two rows because the LiDAR trajectory and LiDAR noise are unchanged; only the IMU noise used by the estimators differs.
\end{flushleft}
\end{table}

\subsection{Localization and Ablation}

\begin{figure}[htbp]
    \centering
    \includegraphics[width=0.96\linewidth]{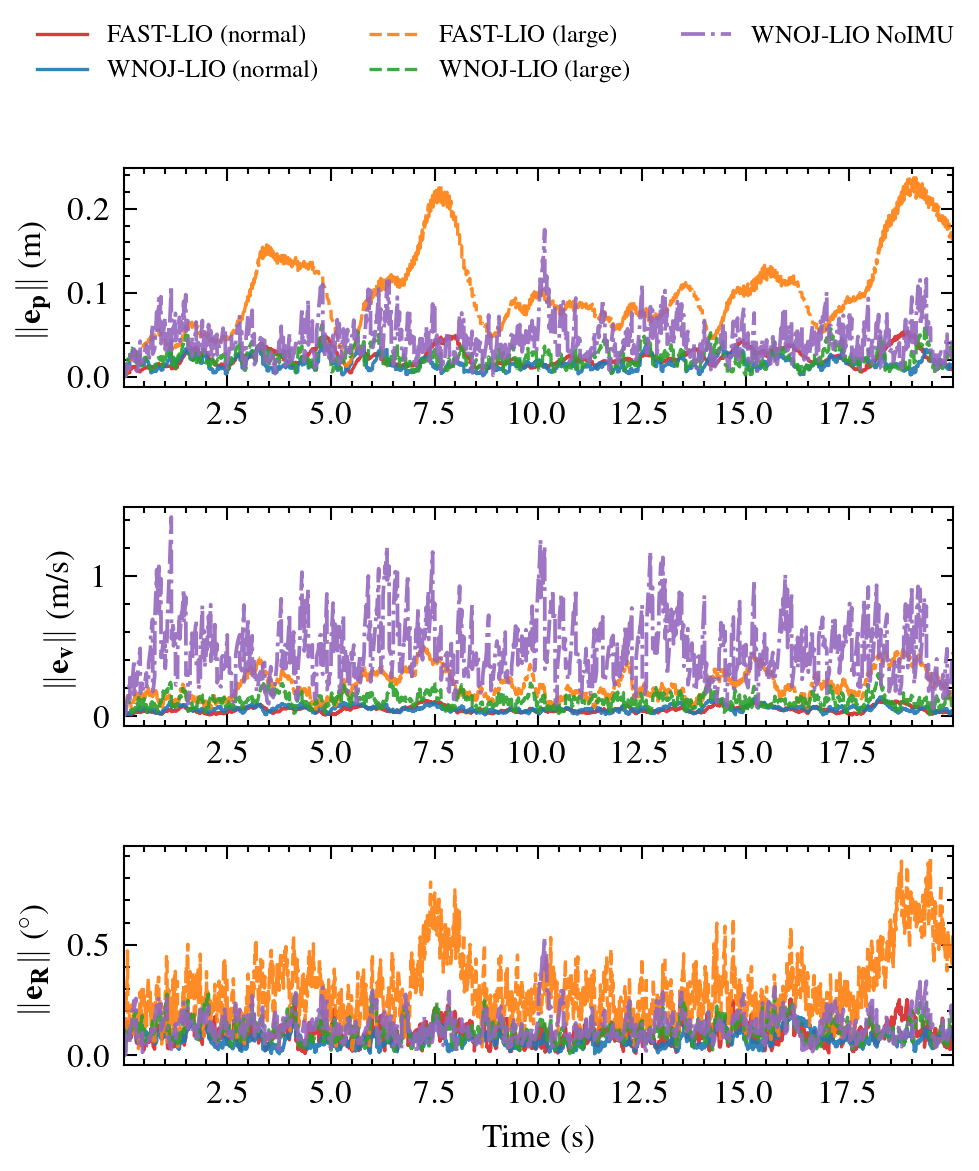}
    \caption{Position, velocity, and attitude error evolution under normal IMU noise, large IMU noise, and the no-IMU ablation. The FAST-LIO-style baseline uses direct IMU propagation, whereas WNOJ-LIO uses WNOJ prediction and treats IMU as measurement updates. Large IMU noise causes pronounced error peaks for the FAST-LIO-style baseline, while WNOJ-LIO remains closer to the normal-noise condition. The no-IMU ablation highlights that LiDAR-only WNOJ updates are insufficient for accurate velocity recovery.}
    \label{fig:sim_noise_pose_error}
\end{figure}

The improved scan de-distortion leads to more accurate point-to-plane registration and consequently improves overall state estimation. Fig.~\ref{fig:sim_noise_pose_error} compares the evolution of position, velocity, and attitude errors for five configurations: FAST-LIO-style and WNOJ-LIO under both normal and high IMU noise, together with a WNOJ-LIO-NoIMU ablation. Under normal IMU noise, both methods achieve accurate localization, while WNOJ-LIO consistently yields lower position and attitude errors with comparable velocity accuracy. As the IMU noise increases, the performance gap becomes significantly larger. Because the FAST-LIO-style baseline directly integrates raw inertial measurements during prediction, inertial disturbances are propagated into both the predicted trajectory and the subsequent scan de-distortion, leading to pronounced error peaks. In contrast, WNOJ-LIO employs the WNOJ prior as the prediction model and incorporates IMU measurements only through recursive measurement updates, making the estimated trajectory substantially less sensitive to inertial noise.

Table~\ref{tab:sim_noise_ablation} quantitatively confirms these observations. Under normal IMU noise, WNOJ-LIO reduces the position RMSE from $0.0257$~m to $0.0176$~m and the attitude RMSE from $0.1098^\circ$ to $0.0802^\circ$, while maintaining comparable velocity accuracy. Under high IMU noise, the advantage becomes much more pronounced, reducing the position RMSE from $0.1193$~m to $0.0247$~m, the velocity RMSE from $0.2421$~m/s to $0.1205$~m/s, and the attitude RMSE from $0.3357^\circ$ to $0.1227^\circ$. These results indicate that decoupling trajectory prediction from raw inertial integration substantially improves robustness against vibration-contaminated IMU measurements.

The WNOJ-LIO-NoIMU ablation further illustrates the complementary roles of the motion prior and inertial measurements. Although the WNOJ prior alone maintains relatively small position and attitude errors by enforcing a smooth physically consistent trajectory, the velocity estimate gradually drifts, resulting in a velocity RMSE of $0.5132$~m/s. This behavior indicates that the motion prior provides strong geometric regularization but cannot fully recover dynamic states without external observations. High-frequency IMU updates therefore remain essential for accurately estimating velocity and other rapidly varying motion quantities.

\begin{table}[htbp]
\caption{Simulation Accuracy Under IMU Noise and Ablation Settings (RMSE)}
\label{tab:sim_noise_ablation}
\centering\footnotesize
\setlength{\tabcolsep}{4pt}
\begin{tabular}{llccc}
\toprule
\multirow{2}{*}{Noise} & \multirow{2}{*}{Method} & Position & Velocity & Attitude \\
 & & (m) & (m/s) & (deg) \\
\midrule
\multirow{2}{*}{Normal} & FAST-LIO-style & 0.0257 & \textbf{0.0523} & 0.1098 \\
 & WNOJ-LIO & \textbf{0.0176} & 0.0572 & \textbf{0.0802} \\
\midrule
\multirow{2}{*}{High} & FAST-LIO-style & 0.1193 & 0.2421 & 0.3357 \\
 & WNOJ-LIO & \textbf{0.0247} & \textbf{0.1205} & \textbf{0.1227} \\
\midrule
Ablation & WNOJ-LIO-NoIMU & 0.0478 & 0.5132 & 0.1447 \\
\bottomrule
\end{tabular}
\begin{flushleft}\vspace{-6pt}
\footnotesize The No-IMU result is a WNOJ-LIO ablation with LiDAR point-to-plane updates only.
\end{flushleft}
\end{table}

\section{Experimental Results}
\label{sec:experiments}

\subsection{Racing-Car Evaluation Protocol}

We evaluated WNOJ-LIO on four high-dynamic racing-car segments collected at Yas Marina Circuit, Abu Dhabi. This dataset is not a public benchmark. It is used here as a challenging real-world test with synchronized LiDAR, raw VectorNav IMU, VectorNav INS/GNSS, Kistler optical velocity, and vibration-isolated Bosch IMU measurements. The selected sequences span a wide range of operating conditions, from low-speed to extreme-speed driving, with maximum vehicle speeds ranging from 53 to 208~km/h. FAST-LIO was adopted as the baseline because it represents the standard LiDAR-inertial design in which the IMU is integrated in the prediction step. The proposed algorithm was implemented in C++ and released at \url{https://github.com/LvJohny/wnoj-ekf-lio.git}. All experiments reported in this paper were evaluated offline on collected data using a desktop computer equipped with an Intel Core i5-11600K CPU.

\begin{figure}[htbp]
    \centering
    \subfloat[]{
        \includegraphics[width=0.35\linewidth]{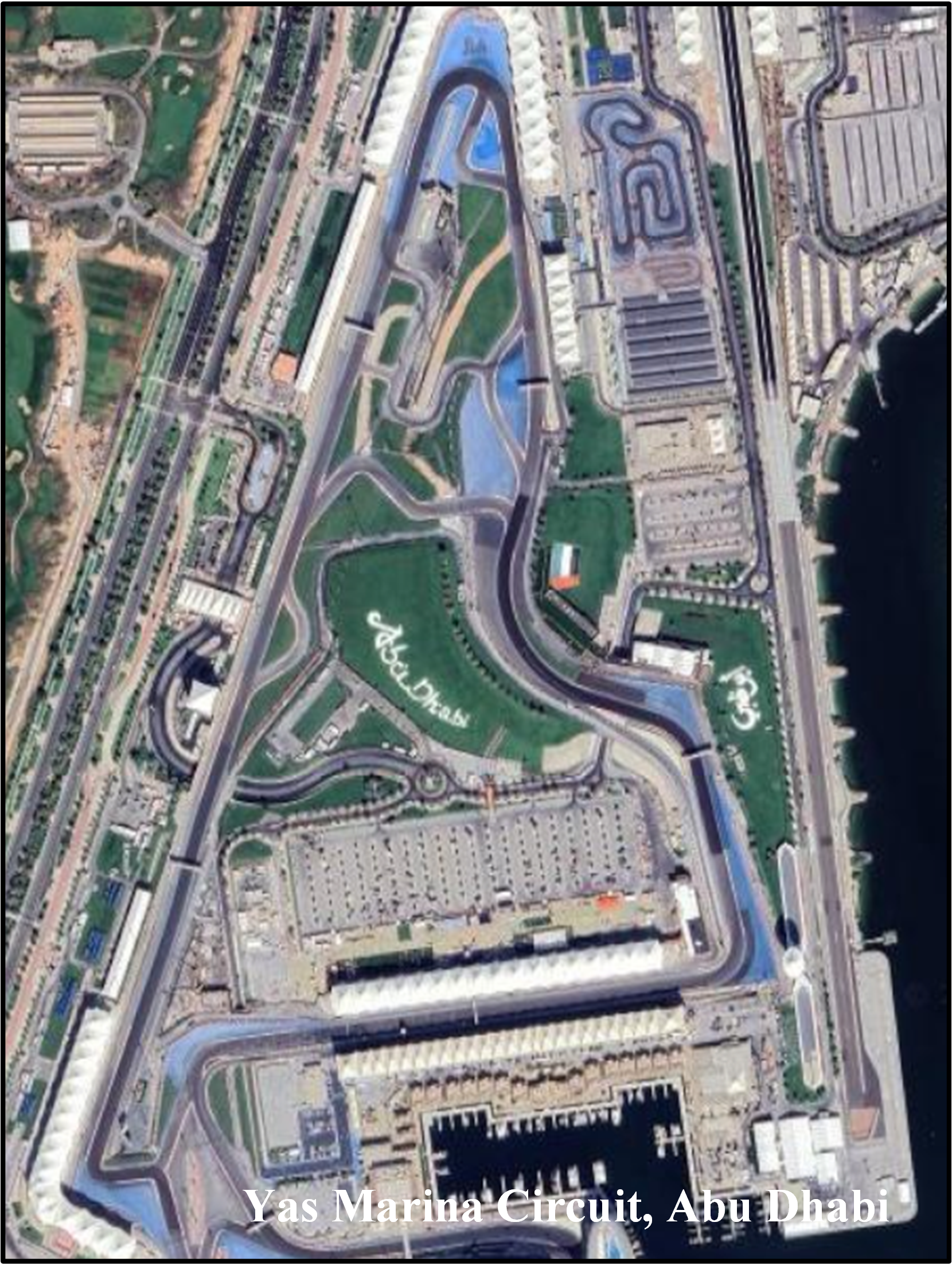}
    }
    \subfloat[]{
        \includegraphics[width=0.58\linewidth]{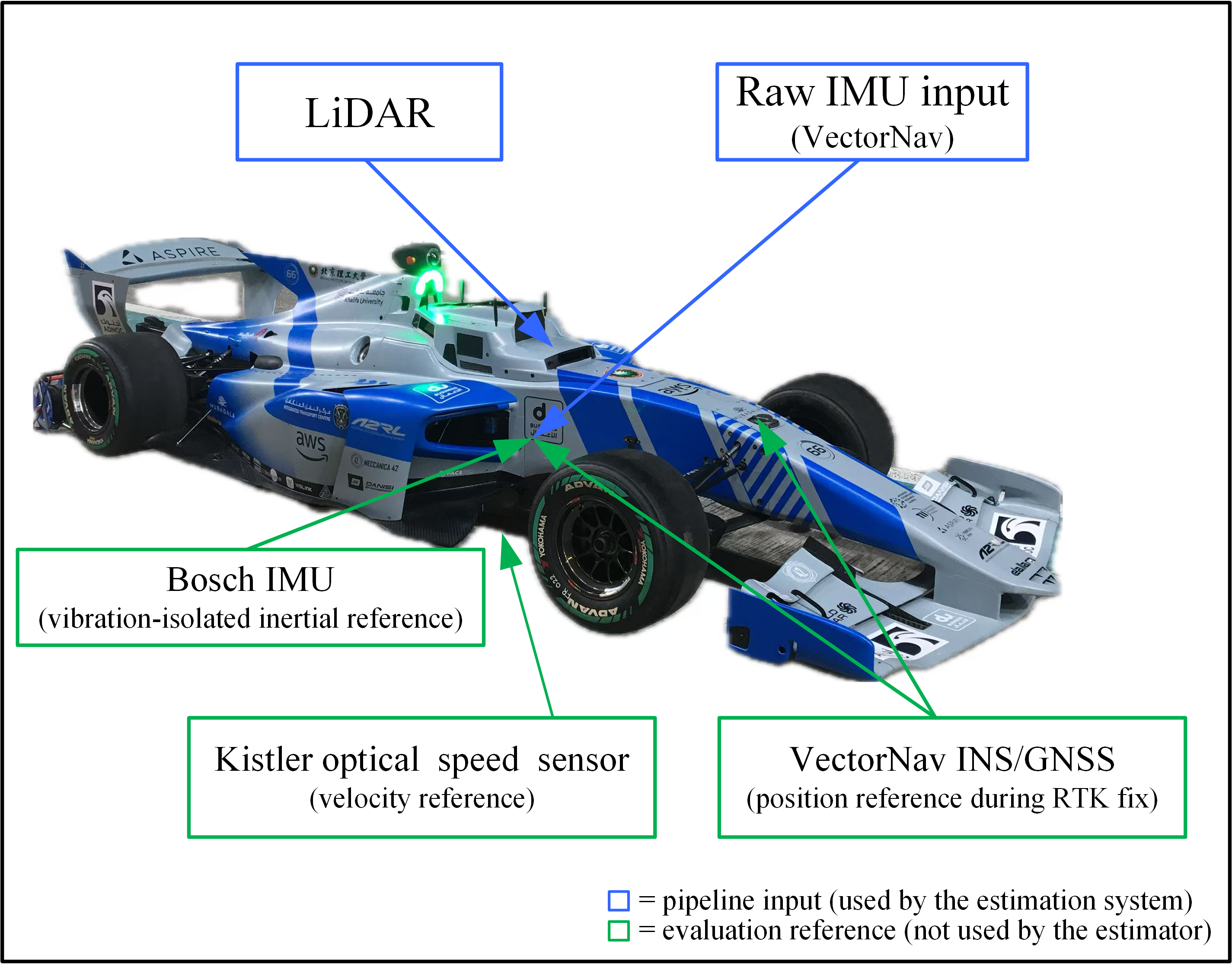}
    }
    \caption{Racing-car experimental site and sensor configuration. (a) Yas Marina Circuit test site. (b) Racing car and sensor configuration. Blue labels denote estimator inputs, including the LiDAR and raw VectorNav IMU. Green labels denote evaluation references: the Kistler optical speed sensor provides body-frame velocity reference, the vibration-isolated Bosch IMU provides an inertial-signal reference, and the VectorNav INS/GNSS provides position reference at RTK-fixed epochs and attitude reference after initial alignment.}
    \label{fig:site_and_car}
\end{figure}

The reference sensors have different roles and reliability properties. The VectorNav INS/GNSS position is used as a position reference only when the RTK solution is fixed and converged. Position RMSE is therefore computed only at RTK-fixed epochs that overlap the LiDAR odometry timestamps within 0.1~s. Velocity is evaluated in the vehicle body frame against the Kistler optical speed sensor, using the forward and lateral velocity components. Attitude is evaluated against the VectorNav INS attitude after a static initial alignment between the local LiDAR-inertial frame and the ENU frame. The vibration-isolated Bosch IMU is not used by either LiDAR-inertial estimator; it is used only for the inertial-signal evaluation. This protocol avoids treating any single sensor as universal ground truth across all metrics.

\begin{figure*}[hbtp]
    \centering
    \subfloat[]{\includegraphics[width=0.4\textwidth]{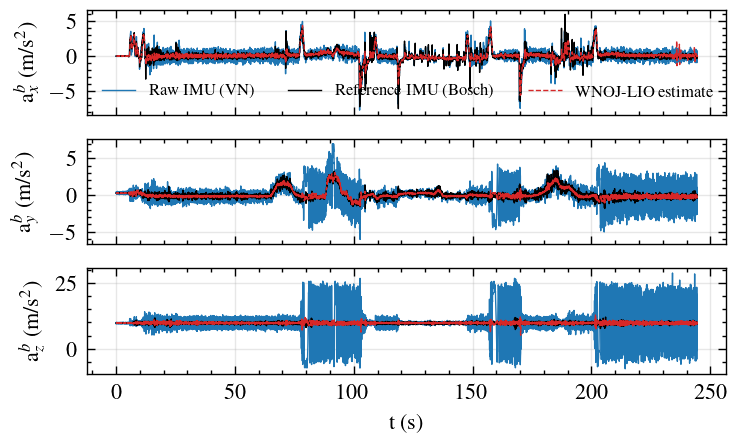}}
    \hfil
    \subfloat[]{\includegraphics[width=0.4\textwidth]{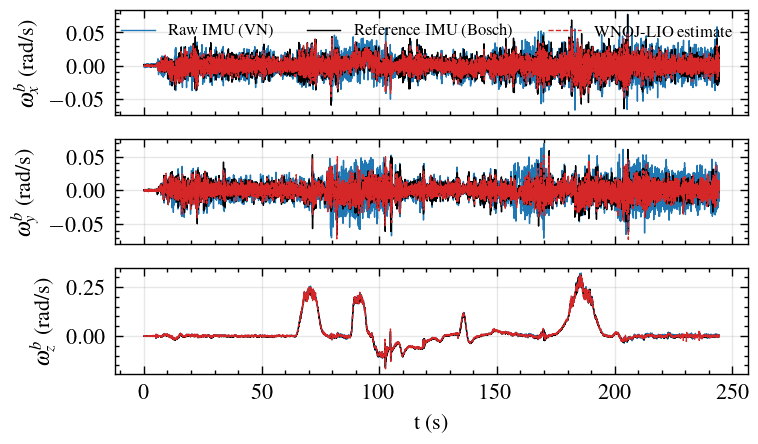}}\\[-1mm]
    \subfloat[]{\includegraphics[width=0.4\textwidth]{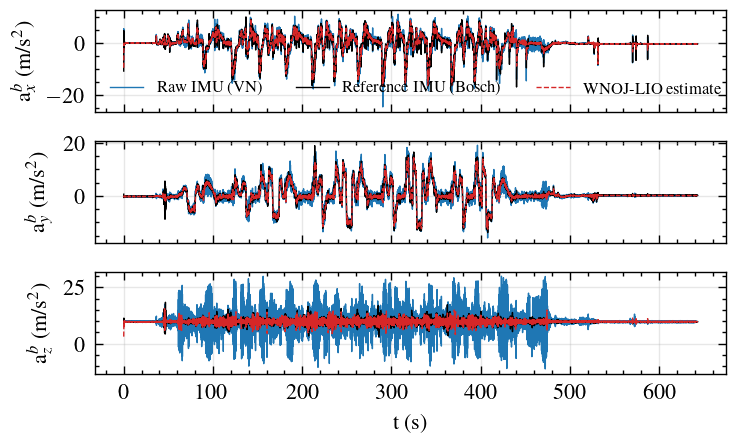}}
    \hfil
    \subfloat[]{\includegraphics[width=0.4\textwidth]{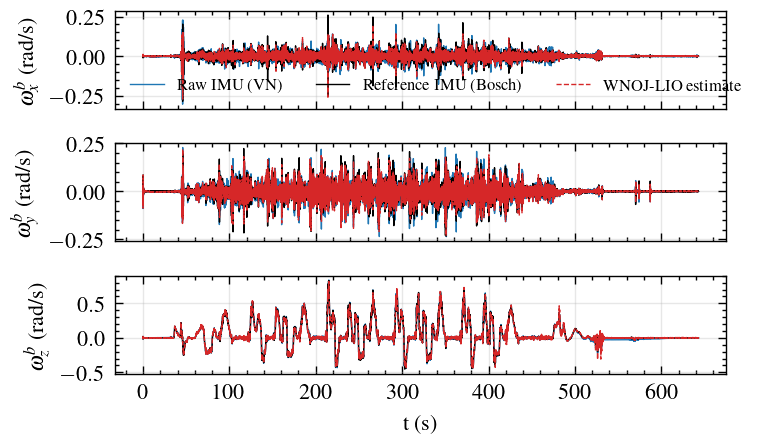}}
    \caption{Inertial signal comparison on representative low-speed and extreme-speed racing-car segments. (a) Acceleration on loop26\_loc at 53~km/h. (b) Angular velocity on loop26\_loc at 53~km/h. (c) Acceleration on loop21\_loc at 208~km/h. (d) Angular velocity on loop21\_loc at 208~km/h. The raw IMU input contains strong vibration-induced components, especially in acceleration, whereas the WNOJ-LIO estimate follows the vibration-isolated Bosch IMU reference more closely.}
    \label{fig:real_imu_signal_comparison}
\end{figure*}

\begin{figure*}[htbp]
    \centering
    \subfloat[]{\includegraphics[width=0.4\textwidth]{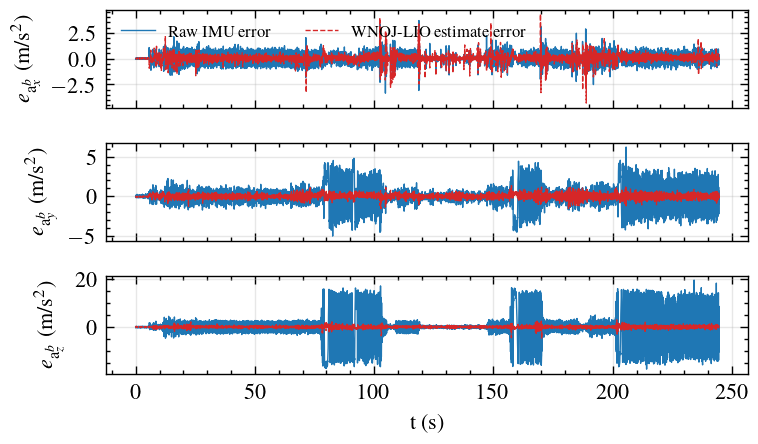}}
    \hfil
    \subfloat[]{\includegraphics[width=0.4\textwidth]{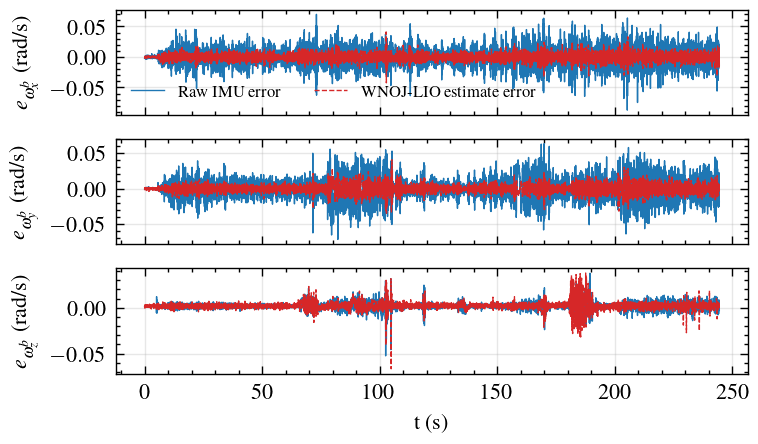}}\\[-1mm]
    \subfloat[]{\includegraphics[width=0.4\textwidth]{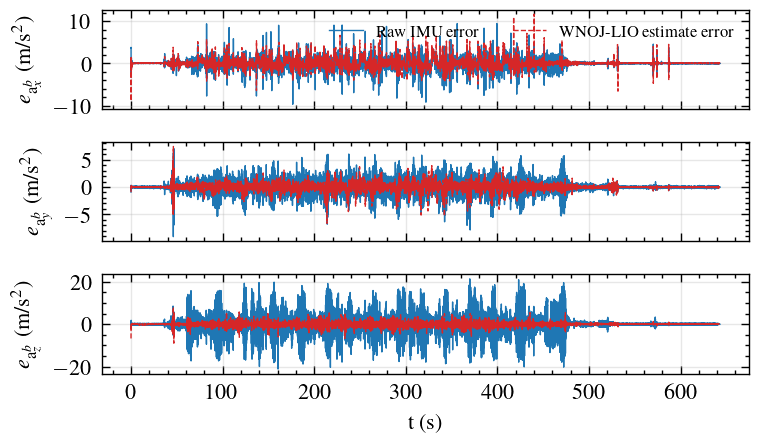}}
    \hfil
    \subfloat[]{\includegraphics[width=0.4\textwidth]{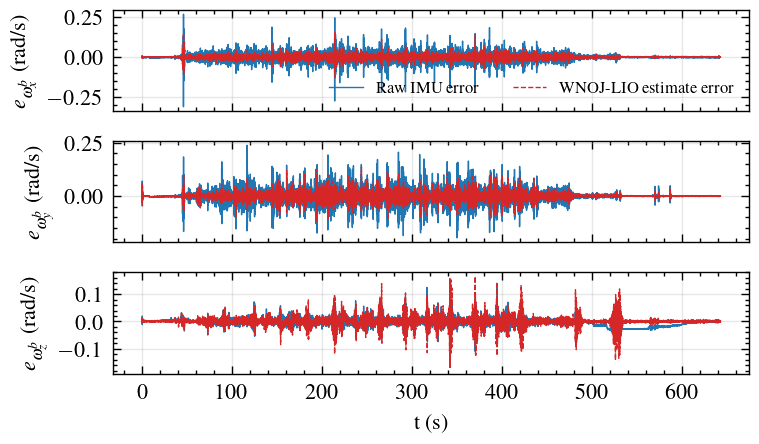}}
    \caption{Inertial signal errors relative to the calibrated Bosch IMU reference. (a) Acceleration error on loop26\_loc at 53~km/h. (b) Angular-velocity error on loop26\_loc at 53~km/h. (c) Acceleration error on loop21\_loc at 208~km/h. (d) Angular-velocity error on loop21\_loc at 208~km/h. The comparison isolates inertial-denoising behavior from the intermittent INS/RTK position reference used later for localization evaluation.}
    \label{fig:real_imu_signal_error}
\end{figure*}

\subsection{Inertial Signal Evaluation}

Before evaluating pose, velocity, and attitude, we directly examined the inertial signals used by the two LiDAR-inertial pipelines. The racing car carried a high-noise VectorNav IMU stream, denoted as the raw IMU input, and a vibration-isolated Bosch IMU, which was used only as an inertial-signal reference for this comparison. The Bosch measurements were calibrated into the VectorNav/vehicle body frame. Both FAST-LIO and WNOJ-LIO receive the same raw IMU input. FAST-LIO directly integrates this signal in prediction. For WNOJ-LIO, the compared angular velocity is the posterior body-frame angular velocity, while the compared acceleration signal is reconstructed by combining the estimated vehicle acceleration with the estimated gravity term and transforming it into the body frame. This comparison evaluates whether the WNOJ prior and high-rate IMU measurement update suppress vibration-induced inertial noise in the signal domain.

\begin{table*}[htbp]
\caption{Inertial Signal RMSE Against Bosch IMU Reference}
\label{tab:real_imu_rmse}
\centering\footnotesize
\setlength{\tabcolsep}{3pt}
\begin{tabular}{lllcccccccc}
\toprule
\multirow{3}{*}{Segment} & \multirow{3}{*}{Max speed} & \multirow{3}{*}{Signal} &
\multicolumn{4}{c}{$\mathbf{a}^b$ RMSE (m/s$^2$)} &
\multicolumn{4}{c}{$\bm{\omega}^b$ RMSE (rad/s)} \\
\cmidrule(lr){4-7}\cmidrule(lr){8-11}
 & & & $x$ & $y$ & $z$ & Total & $x$ & $y$ & $z$ & Total \\
\midrule
\multirow{2}{*}{loop26} & \multirow{2}{*}{53~km/h}
 & Raw IMU & 0.59 & 1.43 & 6.25 & 6.44 & 0.017 & 0.016 & 0.005 & 0.024 \\
 & & WNOJ-LIO estimate & 0.40 & 0.31 & 0.49 & \textbf{0.71} & 0.007 & 0.007 & 0.005 & \textbf{0.011} \\
\midrule
\multirow{2}{*}{loop20} & \multirow{2}{*}{112~km/h}
 & Raw IMU & 0.69 & 1.44 & 6.04 & 6.25 & 0.021 & 0.021 & 0.006 & 0.030 \\
 & & WNOJ-LIO estimate & 0.52 & 0.42 & 0.69 & \textbf{0.96} & 0.012 & 0.010 & 0.007 & \textbf{0.017} \\
\midrule
\multirow{2}{*}{loop38} & \multirow{2}{*}{152~km/h}
 & Raw IMU & 1.29 & 1.45 & 4.72 & 5.10 & 0.035 & 0.028 & 0.015 & 0.047 \\
 & & WNOJ-LIO estimate & 0.89 & 0.77 & 0.86 & \textbf{1.46} & 0.015 & 0.015 & 0.032 & \textbf{0.038} \\
\midrule
\multirow{2}{*}{loop21} & \multirow{2}{*}{208~km/h}
 & Raw IMU & 1.20 & 1.23 & 4.78 & 5.08 & 0.024 & 0.030 & 0.014 & 0.041 \\
 & & WNOJ-LIO estimate & 0.89 & 0.70 & 0.93 & \textbf{1.47} & 0.012 & 0.017 & 0.017 & \textbf{0.027} \\
\bottomrule
\end{tabular}
\begin{flushleft}\vspace{-6pt}
\footnotesize Max speed is the post-start maximum speed of each segment. ``Raw IMU'' denotes the high-noise VectorNav IMU stream provided to both pipelines. For WNOJ-LIO, angular velocity is the posterior body-frame state, and acceleration is reconstructed from the estimated vehicle acceleration and gravity term in the body frame before comparison with the Bosch reference.
\end{flushleft}
\end{table*}

Table~\ref{tab:real_imu_rmse} quantifies the trend shown in Figs.~\ref{fig:real_imu_signal_comparison} and~\ref{fig:real_imu_signal_error}. Across the four selected segments, WNOJ-LIO reduced total acceleration RMSE from 5.08-6.44~m/s$^2$ to 0.71-1.47~m/s$^2$. It also reduced total angular velocity RMSE from 0.024-0.047~rad/s to 0.011-0.038~rad/s. The largest improvement appears in acceleration, where the raw VectorNav signal contains strong vibration-induced components. The angular-velocity results show a smaller but consistent total-RMSE reduction across all segments, although individual axes can trade off under severe high-speed motion.

\subsection{Position, Velocity, and Attitude Accuracy}

\begin{figure}[htbp]
    \centering
    \subfloat[]{\includegraphics[width=\columnwidth]{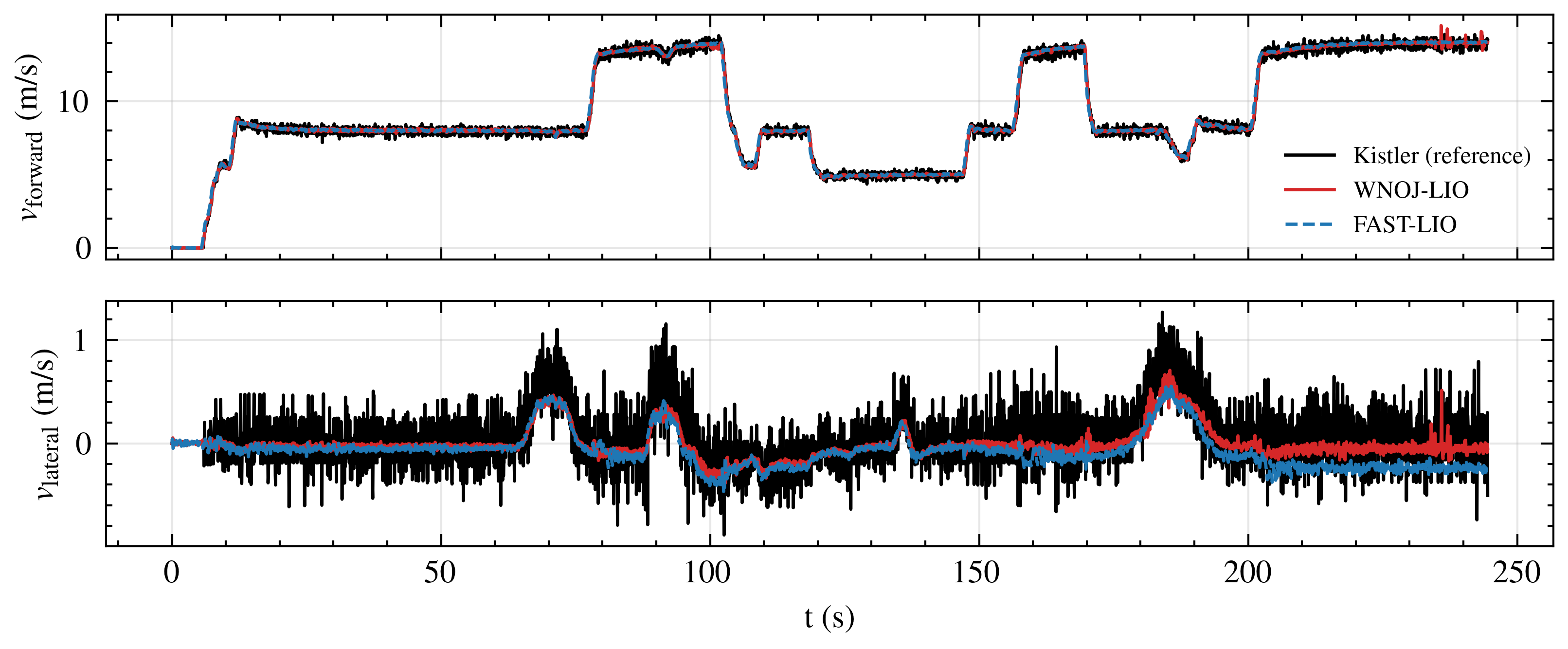}} \\
    \subfloat[]{\includegraphics[width=\columnwidth]{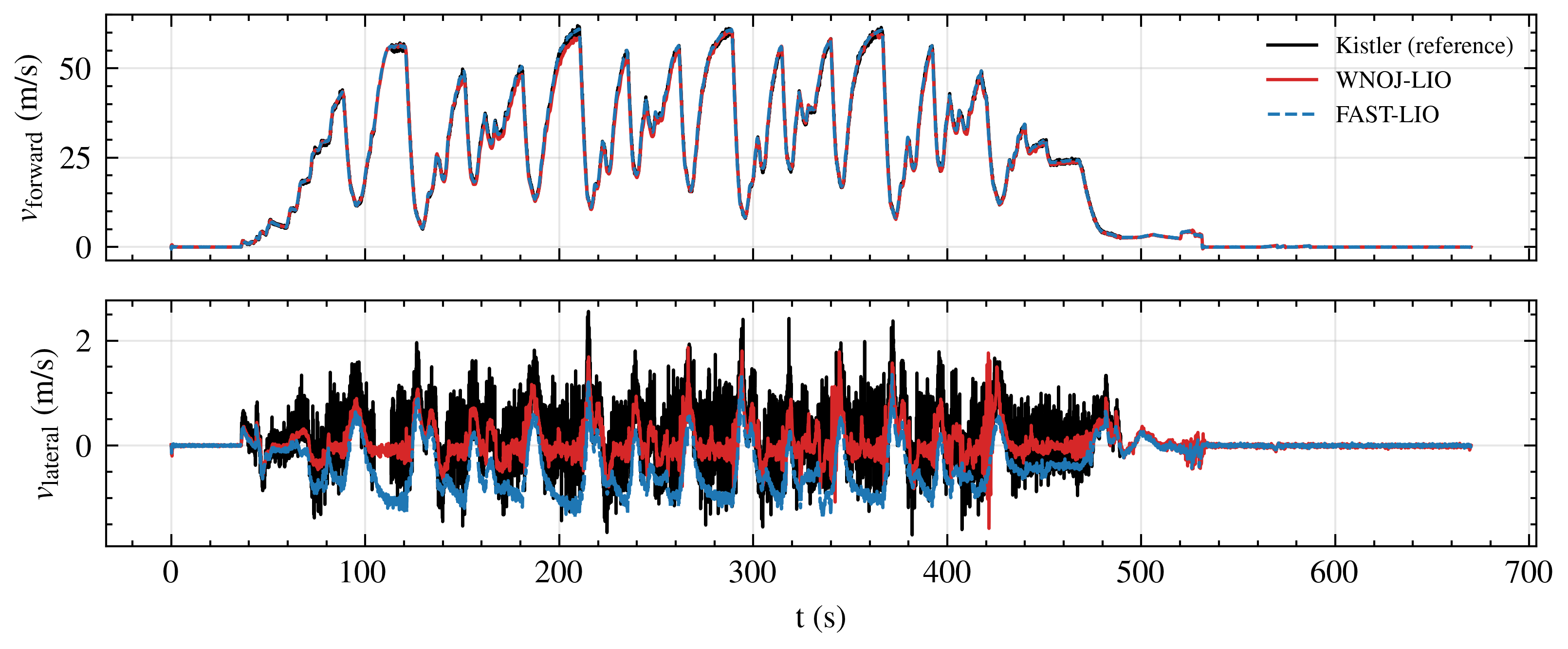}}
    \caption{Velocity comparisons against the Kistler reference on representative racing-car segments. (a) loop26\_loc at 53~km/h. (b) loop21\_loc at 208~km/h. Each comparison contains the forward and lateral body-frame velocity components.}
    \label{fig:low_and_extreme_high_velocity_compare}
\end{figure}

\begin{figure}[htbp]
    \centering
    \subfloat[]{\includegraphics[width=\columnwidth]{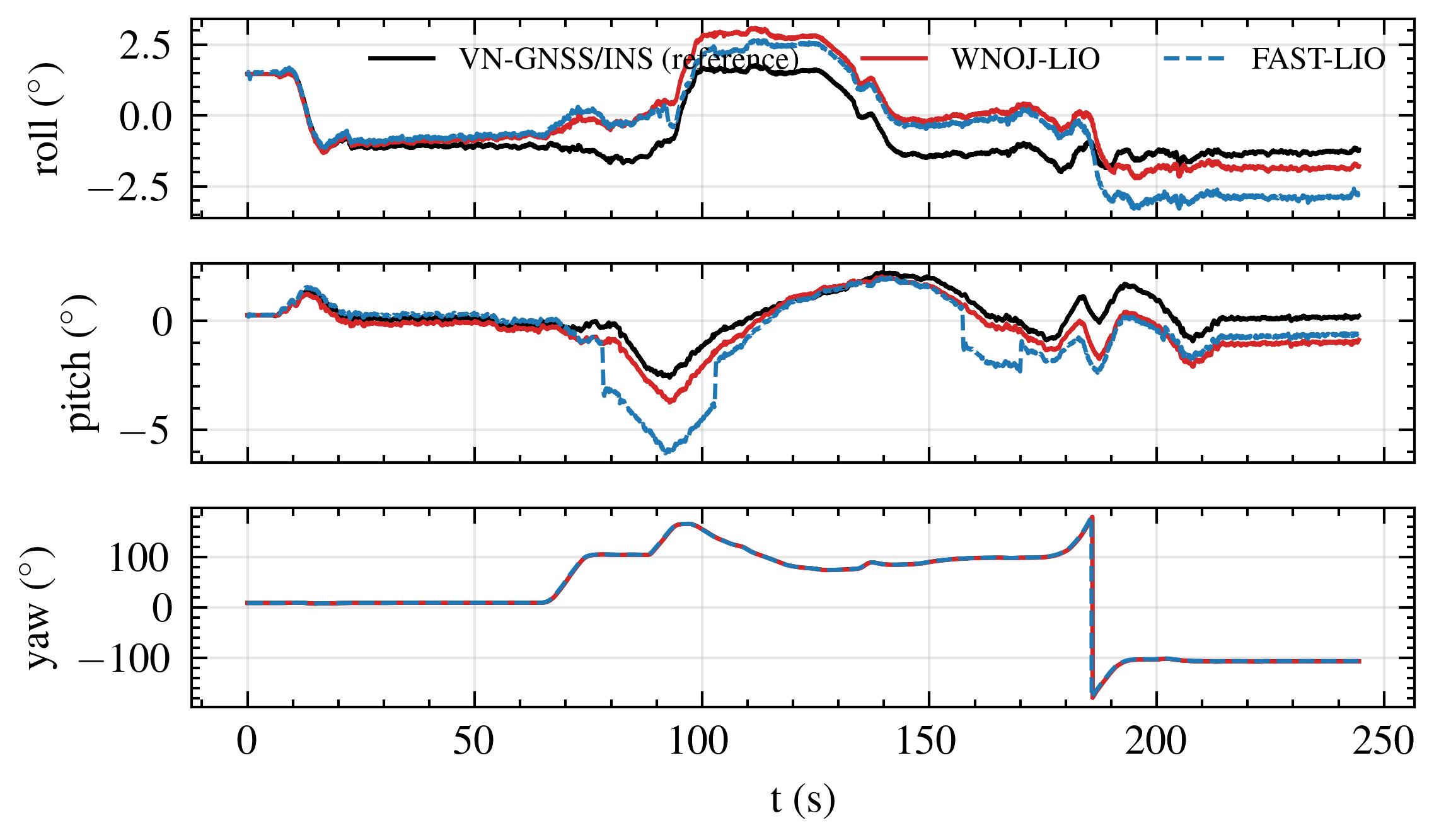}} \\
    \subfloat[]{\includegraphics[width=\columnwidth]{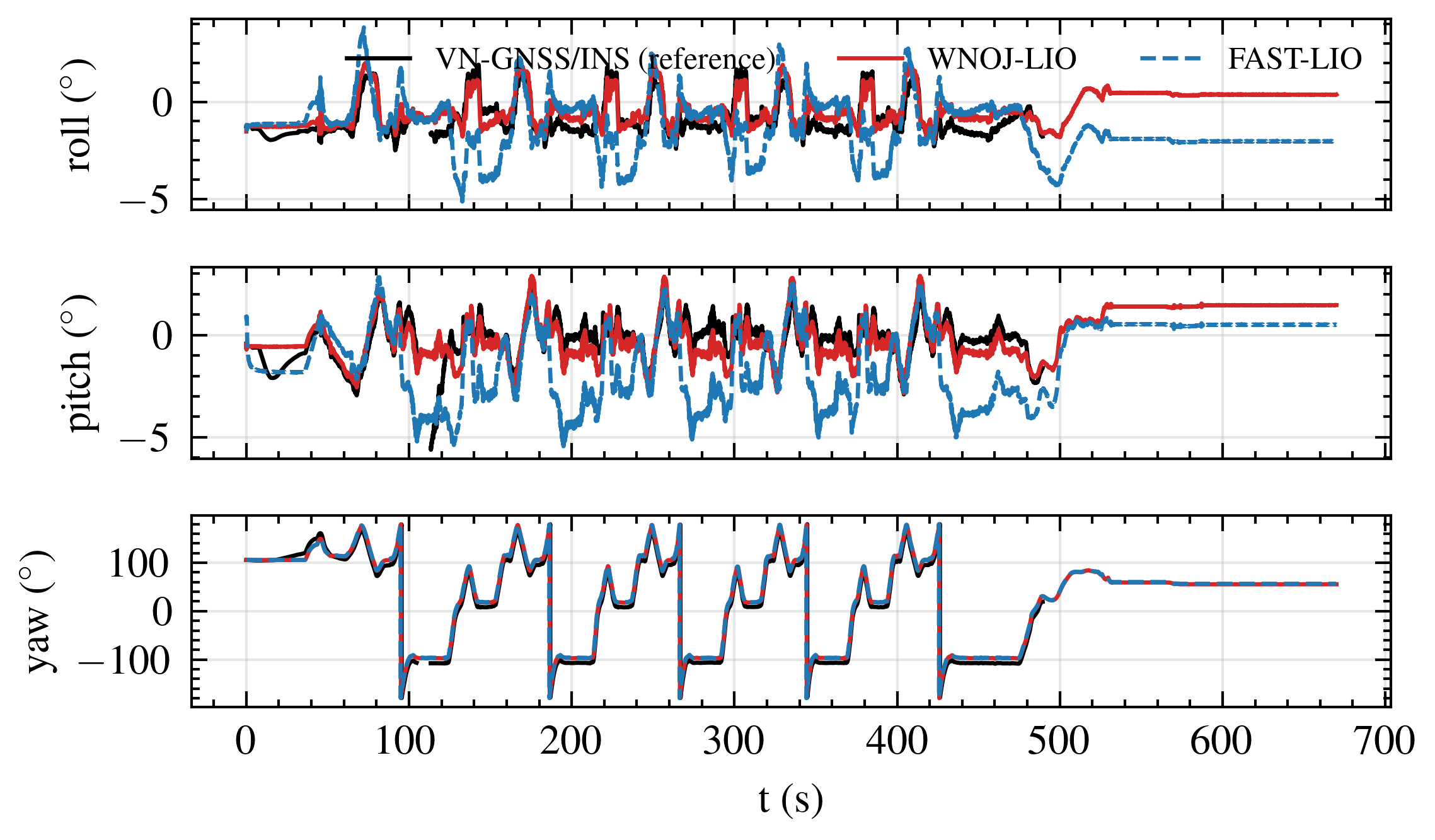}}
    \caption{Attitude comparisons against the VectorNav INS reference on representative racing-car segments. (a) loop26\_loc at 53~km/h. (b) loop21\_loc at 208~km/h. The comparison uses the static initial alignment described in the evaluation protocol.}
    \label{fig:low_and_extreme_high_attitude_compare}
\end{figure}

\begin{figure}[htbp]
    \centering
    \subfloat[]{\includegraphics[width=0.48\linewidth]{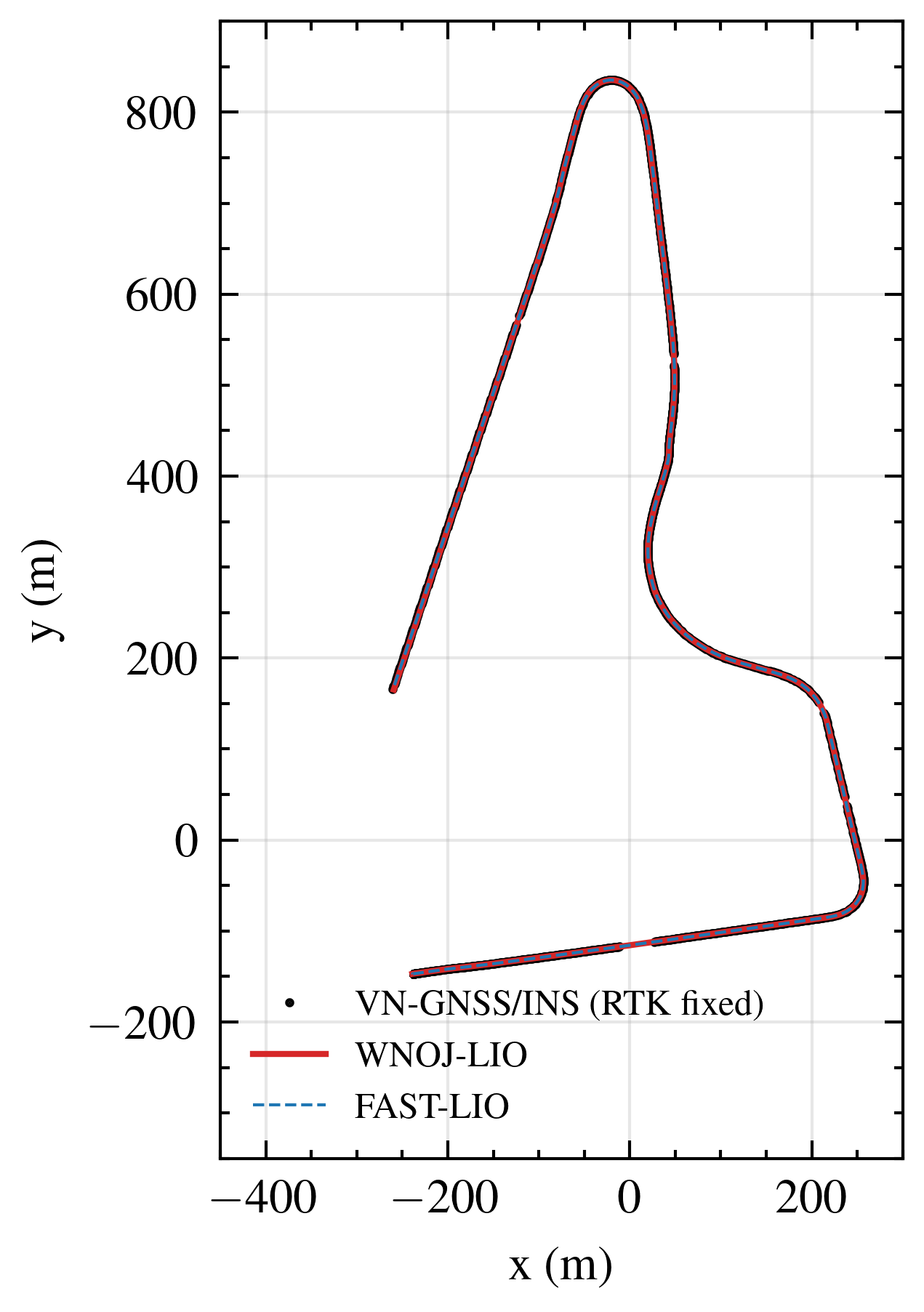}}
    \subfloat[]{\includegraphics[width=0.48\linewidth]{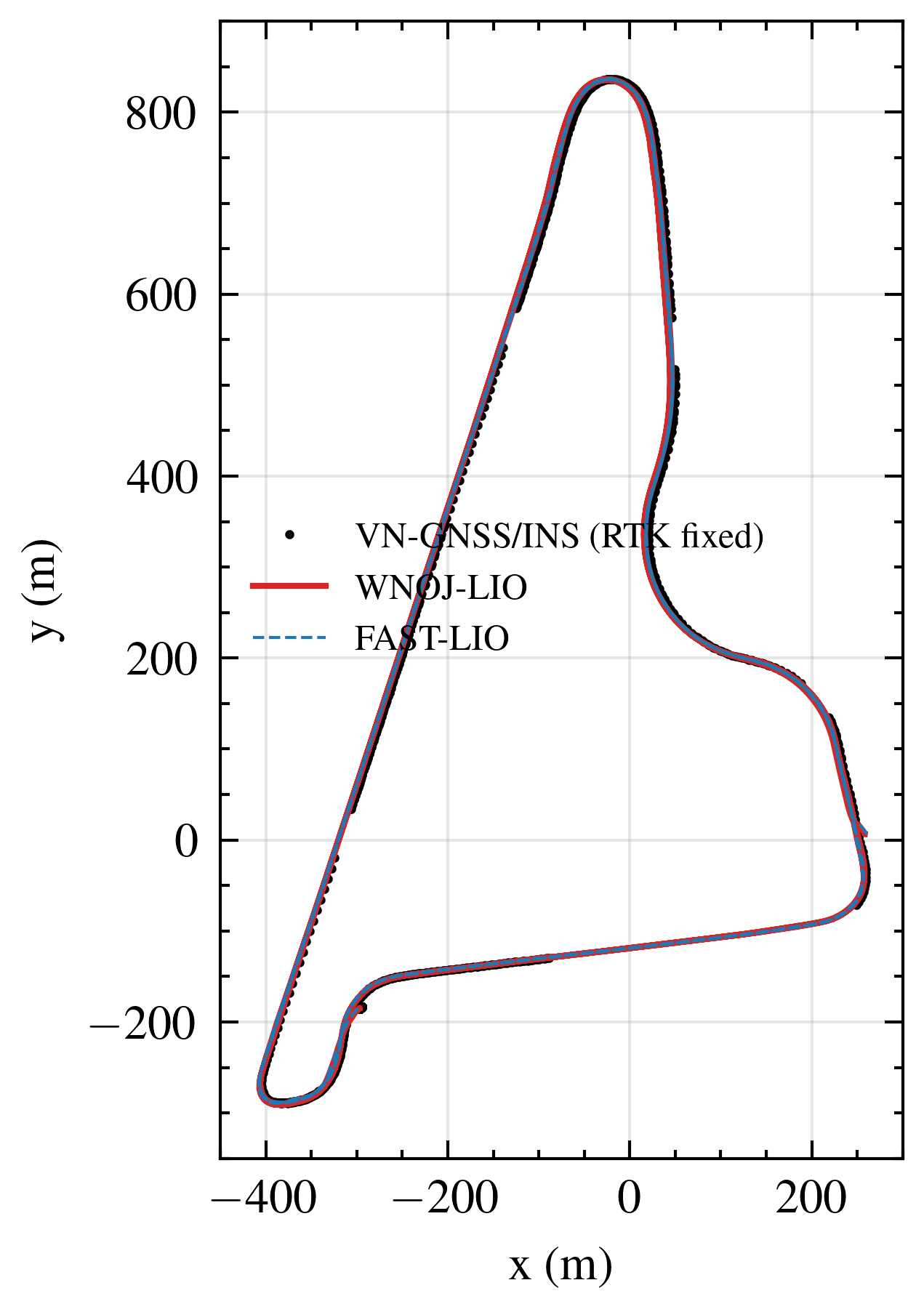}}
    \\
    \subfloat[]{\includegraphics[width=0.96\linewidth]{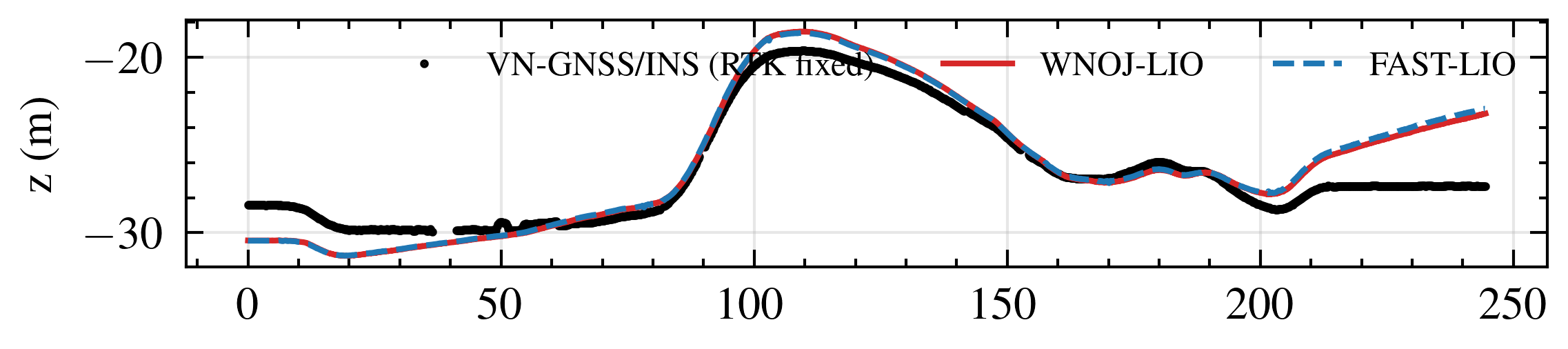}} \\
    \subfloat[]{\includegraphics[width=0.96\linewidth]{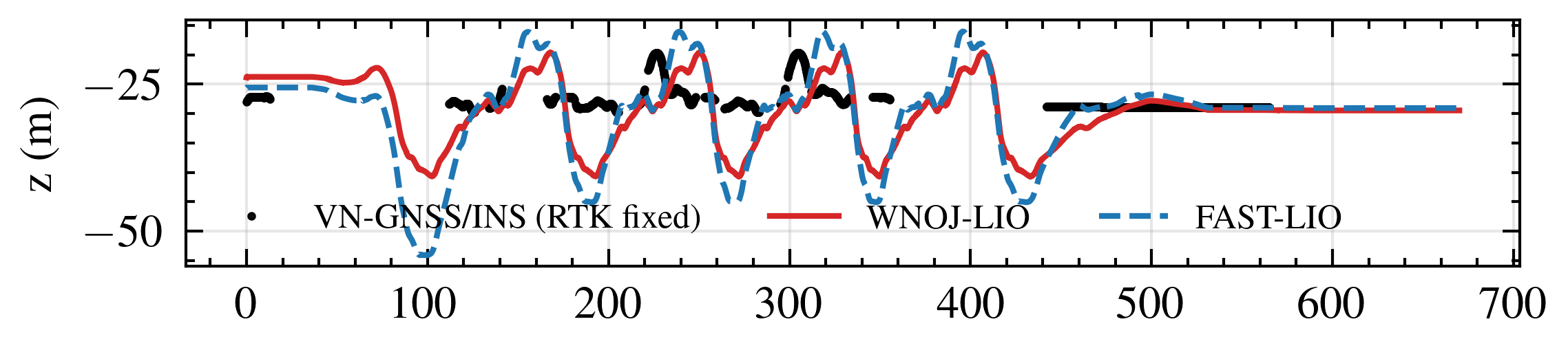}}
    \\
    \caption{Position comparisons against the VectorNav INS/GNSS reference at RTK-fixed epochs. (a) Horizontal trajectory comparison for loop26\_loc at 53~km/h. (b) Horizontal trajectory comparison for loop21\_loc at 208~km/h. (c) Altitude comparison for loop26\_loc at 53~km/h. (d) Altitude comparison for loop21\_loc at 208~km/h.}
    \label{fig:low_and_extreme_high_position_compare}
\end{figure}

Figs.~\ref{fig:low_and_extreme_high_velocity_compare}-\ref{fig:low_and_extreme_high_position_compare} show representative localization results on the low-speed loop26\_loc segment and the extreme-speed loop21\_loc segment. The Kistler velocity plots show that both methods track the dominant forward-speed profile, while WNOJ-LIO is closer to the lateral-velocity reference in the high-speed segment. The attitude plots show larger roll and pitch deviations for FAST-LIO under the repeated high-speed maneuvers in loop21\_loc. The position plots are shown only at RTK-fixed epochs and should therefore be interpreted as intermittent reference checks rather than continuous ground truth.

\begin{table*}[htbp]
\caption{Real-Data Accuracy and Runtime on Racing-Car Segments}
\label{tab:racing_accuracy}
\centering\footnotesize
\setlength{\tabcolsep}{3pt}
\begin{tabular}{lcccccccc}
\toprule
\multirow{2}{*}{Dataset} & \multirow{2}{*}{RTK fixed / total} & \multicolumn{2}{c}{Position RMSE (m)} & \multicolumn{2}{c}{Kistler velocity RMSE (m/s)} & \multicolumn{2}{c}{Attitude RMSE (deg)} & Runtime (ms/frame) \\
\cmidrule(lr){3-4}\cmidrule(lr){5-6}\cmidrule(lr){7-8}
 & & WNOJ-LIO & FAST-LIO & WNOJ-LIO & FAST-LIO & WNOJ-LIO & FAST-LIO & WNOJ / FAST \\
\midrule
loop26\_loc & 1180 / 2998 & \textbf{1.51} & 1.57 & \textbf{0.313} & 0.336 & \textbf{1.20} & 1.76 & 44.15 / \textbf{20.21} \\
loop20\_loc & 2770 / 3688 & \textbf{8.88} & 14.82 & \textbf{0.398} & 0.442 & \textbf{1.71} & 3.10 & 45.98 / \textbf{19.66} \\
loop38\_loc & 1380 / 3612 & 6.87 & \textbf{6.68} & \textbf{1.028} & 1.197 & \textbf{29.31} & 31.13 & 45.75 / \textbf{22.37} \\
loop21\_loc & 1449 / 2341 & \textbf{7.84} & 8.53 & \textbf{0.887} & 0.940 & \textbf{9.84} & 10.92 & 43.78 / \textbf{18.47} \\
\bottomrule
\end{tabular}
\begin{flushleft}\vspace{-6pt}
\footnotesize ``RTK fixed / total'' reports the number of RTK-fixed samples used for position RMSE and the total number of matched odometry samples. Position RMSE is computed only at RTK-fixed epochs with timestamp mismatch no greater than 0.1~s. Velocity RMSE is computed against Kistler forward/lateral velocity in the body frame. Attitude RMSE is computed against VectorNav INS attitude after static initial alignment of each estimator's local frame to ENU.
\end{flushleft}
\end{table*}

Table~\ref{tab:racing_accuracy} summarizes all four racing-car segments. WNOJ-LIO achieved lower RTK-fixed position RMSE on loop26\_loc, loop20\_loc, and loop21\_loc, while FAST-LIO was slightly lower only on loop38\_loc. This position result should still be interpreted with the intermittent RTK-fixed GNSS reference and the vertical sensitivity seen in Fig.~\ref{fig:low_and_extreme_high_position_compare}. In contrast, WNOJ-LIO achieved lower Kistler-referenced body-frame velocity RMSE and lower INS-referenced attitude RMSE on all four segments. The loop21\_loc segment also shows the highest WNOJ-LIO runtime, with an average processing time of 78.63~ms/frame. These metrics are available over broader intervals than the RTK-fixed position samples and therefore provide complementary evidence for high-dynamic vehicle motion estimation.

\subsection{Map Quality Observations}

\begin{figure*}[htbp]
    \centering
    \subfloat[]{\includegraphics[width=0.95\textwidth]{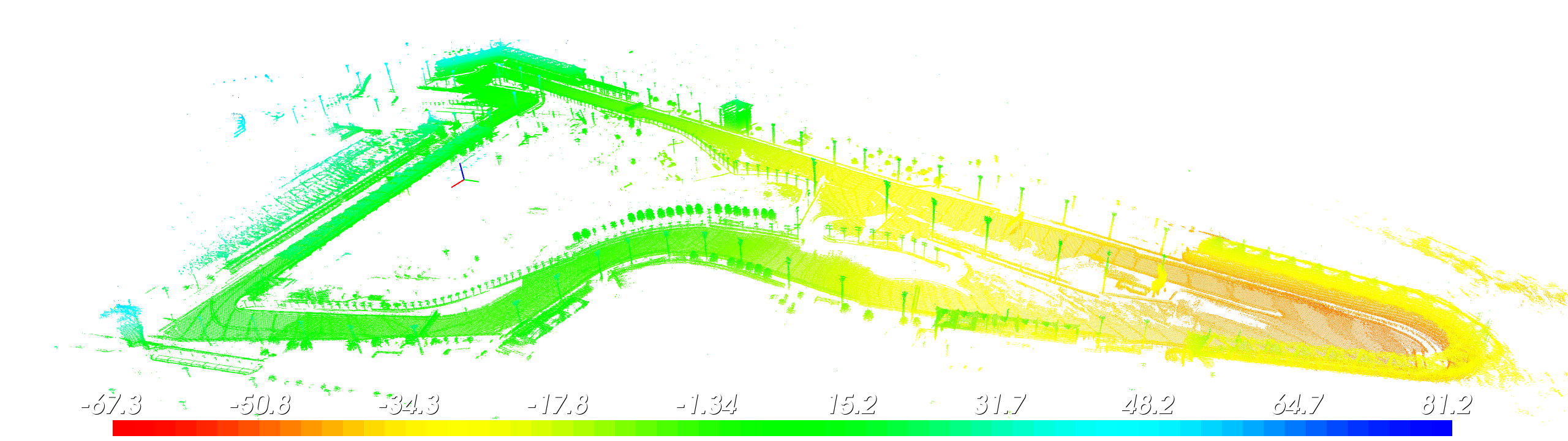}}\\[-1mm]
    \subfloat[]{\includegraphics[width=0.95\textwidth]{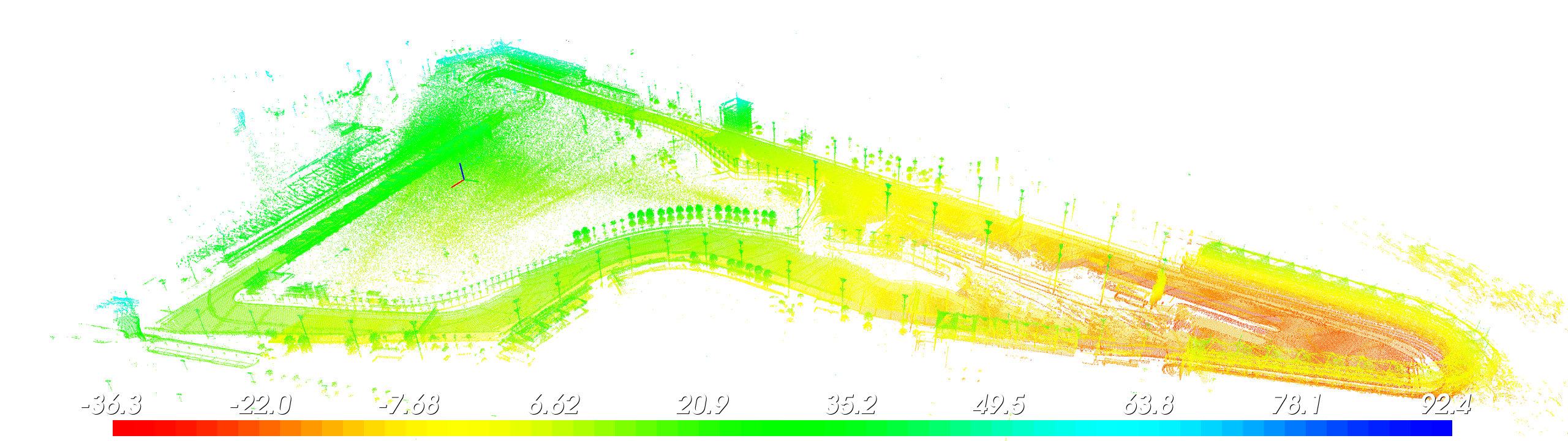}}
    \caption{Qualitative point-cloud map comparison on the extreme-speed loop21\_loc segment. (a) WNOJ-LIO map. (b) FAST-LIO map. Both maps use the same white-background rendering and height-colored point cloud visualization.}
    \label{fig:point_cloud_map_quality}
\end{figure*}

The generated point-cloud maps provide a qualitative check on the scan-to-map consistency that is not fully captured by intermittent RTK samples. Fig.~\ref{fig:point_cloud_map_quality} compares WNOJ-LIO and FAST-LIO on the extreme-speed loop21\_loc segment. Both maps recover the main track geometry, but the FAST-LIO map shows thicker point dispersion and visible ghosting around several structures. The WNOJ-LIO map is more compact in these regions. This observation is used only as qualitative map evidence; the quantitative conclusions remain those reported in Table~\ref{tab:racing_accuracy}.

\section{Conclusion}
\label{sec:conclusion}

This paper presented WNOJ-LIO, a LiDAR-IMU fusion framework for high-dynamic motion estimation. Its dual-frequency architecture uses the IMU as a high-frequency EKF measurement update and LiDAR as a low-frequency geometric update for pose correction. The decoupled WNOJ process model evolves the global state on $\R^3 \times \SO(3)$ and represents attitude perturbations in $\so(3)$, enabling recursive covariance propagation through the global-state transition Jacobian $\bm{\Phi}'$, the local-to-global process-noise mapping $\bm{\Psi}'$, and the closed-form integrated WNOJ covariance $\mathbf{Q}_{t_i}$. This formulation transfers the WNOJ prior from batch trajectory estimation to a real-time recursive estimator. Simulations demonstrated improved acceleration and angular-velocity denoising, point-cloud de-distortion, and localization accuracy compared with a FAST-LIO-style baseline. Racing-car experiments on four segments spanning 53 to 208~km/h further yielded lower RMSE than FAST-LIO in total acceleration, angular velocity, body-frame velocity, and attitude across all four segments. Under the intermittent RTK-fixed reference, WNOJ-LIO maintained competitive position accuracy, yielding lower RMSE in three of the four segments. The qualitative map comparison on the extreme-speed segment also showed reduced point dispersion for WNOJ-LIO. 

Future work will focus on adaptive WNOJ noise tuning, online reference-quality assessment, sliding-window extensions for LiDAR latency, and integration with visual measurements for LiDAR-visual-inertial odometry.

\section*{Acknowledgment}
The authors would like to thank the data-collection team for their support during the racing-car tests.

\bibliographystyle{IEEEtran}

\bibliography{main}

\end{document}